\documentclass[10pt,twocolumn,letterpaper]{article}

\usepackage{iccv}
\usepackage{times}
\usepackage{epsfig}
\usepackage{graphicx}
\usepackage{amsmath}
\usepackage{amssymb}

\usepackage[misc]{ifsym}
\usepackage{adjustbox,diagbox}
\usepackage{caption}
\usepackage{dblfloatfix}

\usepackage[accsupp]{axessibility}  

\usepackage[pagebackref=true,breaklinks=true,letterpaper=true,colorlinks,bookmarks=false]{hyperref}

\iccvfinalcopy 

\def\iccvPaperID{1623} 

\begin{document}

\title{Scenimefy: Learning to Craft Anime Scene via Semi-Supervised\\
Image-to-Image Translation\vspace{-0.3cm}}

\author{Yuxin Jiang$^*$ \hspace{12pt} Liming Jiang$^*$ \hspace{12pt} Shuai Yang \hspace{12pt} Chen Change Loy\\[2pt]
S-Lab, Nanyang Technological University \hspace{12pt} \\[1pt]
{\tt\small \{c200203, liming002, shuai.yang, ccloy\}@ntu.edu.sg} \hspace{12pt}
}

\twocolumn[{%
\renewcommand\twocolumn[1][]{#1}%
\maketitle
\begin{center}
    \vspace{-5mm}
\includegraphics[width=\linewidth]{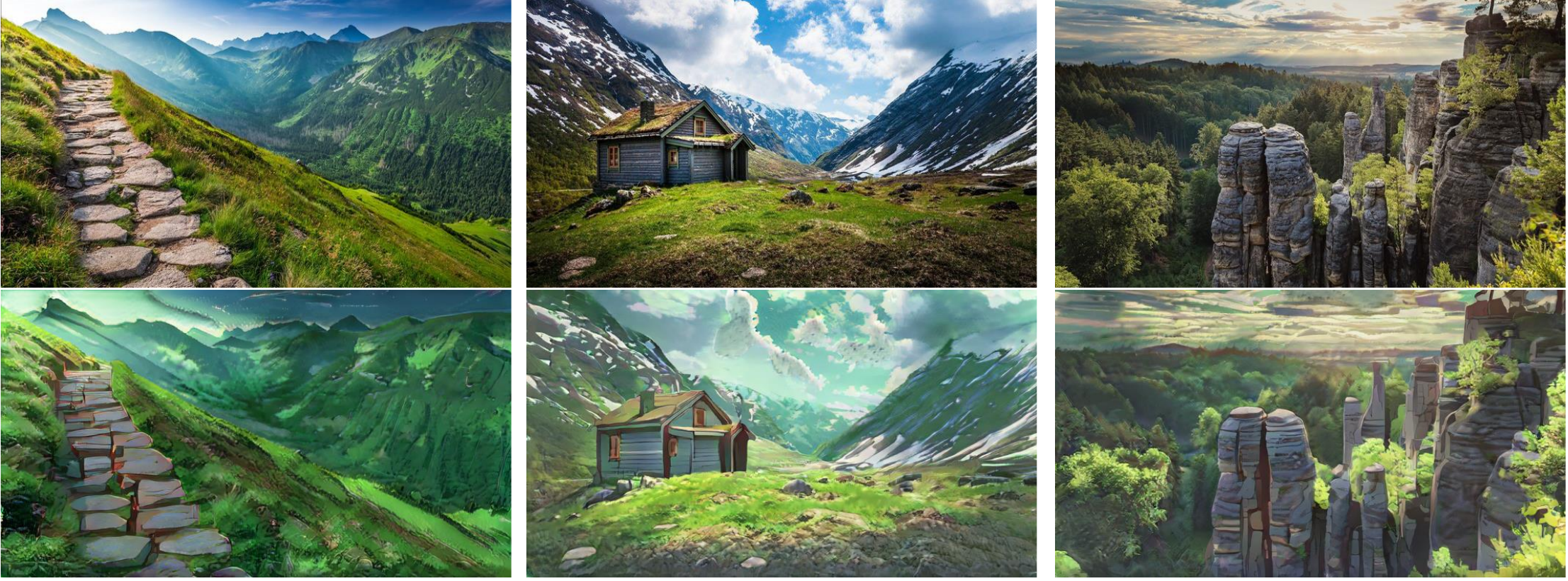}
\vspace{-5mm}
\captionof{figure}{Examples of anime scene rendering by Scenimefy. Top row: input images; Bottom row: our translated results.}
\label{fig:teaser}
\end{center}%
}]

\maketitle

\begin{abstract}
\label{sec:abstract}

Automatic high-quality rendering of anime scenes from complex real-world images is of significant practical value. The challenges of this task lie in the complexity of the scenes, the unique features of anime style, and the lack of high-quality datasets to bridge the domain gap. 
Despite promising attempts, previous efforts are still incompetent in achieving satisfactory results with consistent semantic preservation, evident stylization, and fine details. 
In this study, we propose Scenimefy, a novel semi-supervised image-to-image translation framework that addresses these challenges. Our approach guides the learning with structure-consistent pseudo paired data, simplifying the pure unsupervised setting. 
The pseudo data are derived uniquely from a semantic-constrained StyleGAN leveraging rich model priors like CLIP. We further apply segmentation-guided data selection to obtain high-quality pseudo supervision. 
A patch-wise contrastive style loss is introduced to improve stylization and fine details.
{\let\thefootnote\relax\footnotetext{$^*$ Equal contribution.}}
Besides, we contribute a high-resolution anime scene dataset to facilitate future research. Our extensive experiments demonstrate the superiority of our method over state-of-the-art baselines in terms of both perceptual quality and quantitative performance. {Project page: \url{https://yuxinn-j.github.io/projects/Scenimefy.html}}.

\end{abstract}
\vspace{-4mm}
\section{Introduction}
\label{sec:introduction}

Crafting anime scenes requires significant artistic skill and time, making developing learning-based techniques for automatic scene stylization of unquestionable practical and commercial value. Recent advances in Generative Adversarial Networks (GANs) have led to a significant improvement in automatic stylization, but most research in this area has focused primarily on human faces~\cite{yang2022pastiche, song2021agilegan, men2022dct, li2023parsing, yang2022Vtoonify}. Despite its high research value, generating high-quality anime scenes from complex real-world scene images remains underexplored.

\begin{figure}[t]
    \centering
    \includegraphics[width=\linewidth]
    {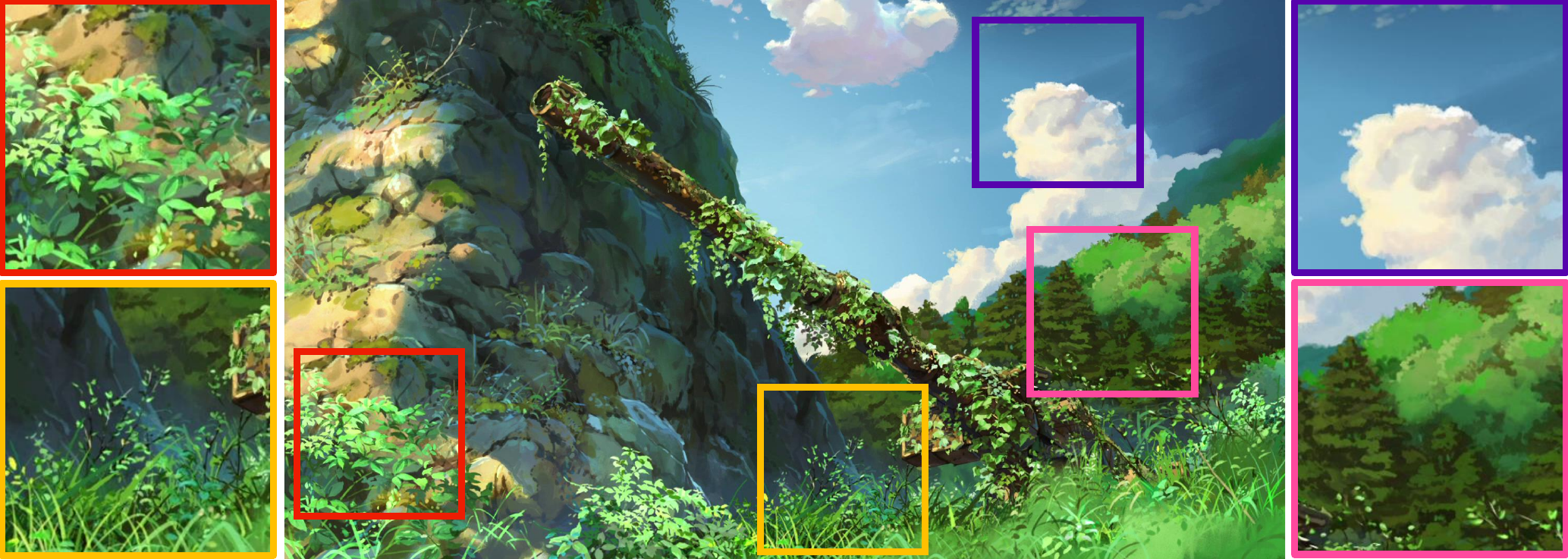}
    \caption{{\bf Characteristics of anime scenes.} A scene frame from Shinkai's film 'Children Who Chase Lost Voices' (2011) shows the presence of hand-drawn brush strokes of grass and stones (foreground), as well as trees and clouds (background), as opposed to clear edges and flat surfaces.}
    \vspace{-2mm}
    \label{fig:animetexture}
\end{figure}

Transferring real scene images into anime styles remains a formidable challenge due to several factors. 
1) \textit{The nature of a scene.}
Scenes are typically composed of multiple objects with complex relationships among them, and there is an inherent hierarchy between foreground and background elements, as shown in Figure~\ref{fig:animetexture}. 
2) \textit{The features of anime.} 
Anime is characterized by unique textures and intricate details, such as the pre-designed brush strokes used in natural landscapes like grass, trees, and clouds, as illustrated in Figure~\ref{fig:animetexture}. These textures are typically organic and hand-drawn, making their style much more difficult to mimic than the sharp edges and smooth color patches defined in previous studies~\cite{chen2018cartoongan, wang2020learning}. 
3) \textit{The domain gap and lack of data.}
There is a large domain gap between real and anime scenes, and a high-quality anime scene dataset is essential in bridging this gap. However, existing datasets contain many human faces and other foreground objects, whose style is different from that of the background scene, leading to their low quality.

Unsupervised image-to-image translation~\cite{zhu2017unpaired, liu2017unsupervised, huang2018munit, jiang2020tsit, park2020contrastive, jung2022exploring} is a typical solution for complex scene stylization without paired training data. 
Despite promising results, existing methods~\cite{chen2018cartoongan, chen2020animegan, wang2020learning, gao2022learning} that focus on anime styles fall short in several ways. 
First, the absence of pixel-wise correspondence in complex scenes hinders existing methods~\cite{chen2018cartoongan, chen2020animegan} from effectively performing evident texture stylization while preserving semantic content, resulting in potentially unnatural results with notable artifacts. 
Second, some approaches~\cite{wang2020learning, gao2022learning} fall short of generating fine details of anime scenes. This is due to their handcrafted anime-specific losses or pre-extracted representations that impose edge and surface smoothness.

To address the challenges discussed above, we propose a novel semi-supervised image-to-image (I2I) translation pipeline, named \textit{Scenimefy}, for producing high-quality anime-style renderings of scene images, as shown in Figure~\ref{fig:teaser}. 
Our key idea is to incorporate a new supervised training branch into the unsupervised framework using generated pseudo paired data to overcome the difficulties of unsupervised training. 
Specifically, we leverage the desirable properties of StyleGAN~\cite{karras2019style, karras2020analyzing} by fine-tuning it to generate coarse paired data between real and anime, which we call pseudo paired data. 
We propose a novel semantic-constrained fine-tuning strategy that leverages rich pre-trained model priors, such as CLIP~\cite{radford2021learning} and VGG~\cite{simonyan2014very}, to guide StyleGAN to capture complex scene features and alleviate overfitting. 
We further introduce a segmentation-guided data selection scheme to filter low-quality data. With the pseudo paired data, Scenimefy learns effective pixel-wise correspondence and generates fine details between the two domains, guided by a novel patch-wise contrastive style loss. 
Together with the unsupervised training branch, our semi-supervised framework seeks a desired trade-off between the faithfulness and fidelity of scene stylization.

To facilitate training, we also collected a high-quality pure anime scene dataset. We conducted comprehensive experiments that demonstrate the effectiveness of Scenimefy, surpassing state-of-the-art baselines in both perceptual quality and quantitative evaluation. In summary, our key contributions are as follows:
\begin{itemize}
\vspace{-0.2cm}
\item We propose a novel semi-supervised image-to-image translation framework for scene stylization that generates high-quality complex anime scene images from real ones. Our framework incorporates a new patch-wise contrastive style loss to improve stylization and fine details.
\vspace{-0.2cm}
\item The training supervision is derived from structure-consistent pseudo paired data generated by a newly designed semantic-constrained StyleGAN fine-tuning strategy with rich pre-trained prior guidance, followed by a segmentation-guided data selection scheme.
\vspace{-0.2cm}
\item We collected a high-resolution anime scene dataset to facilitate future research in scene stylization.
\end{itemize}
\begin{figure*}[!t]
    \centering
    \includegraphics[width=0.99\linewidth]
    {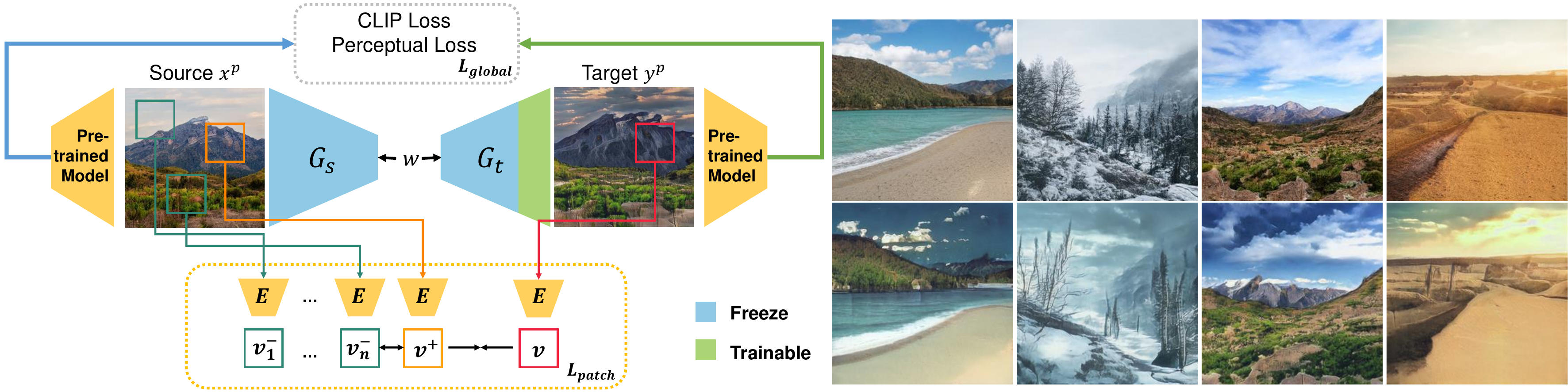}
    \caption{{\bf Overview of our semantic-constrained fine-tuning strategy for pseudo paired data generation.}
    \textbf{Left:} We initialize a source generator $G_s$ and a target one $G_t$ by pre-trained on a real scene domain. $G_s$ remains fixed throughout the process. $G_t$ is optimized using rich pre-trained model prior (\ie, CLIP, VGG) guidance with the early layers freezed to be adapted to an anime scene domain.
    A patch-wise contrastive loss using pre-trained CLIP embedders $E$ is applied to better preserve local spatial details.
    \textbf{Right:} Examples of generated pseudo paired data after segmentation-guided data selection.
    }
    \vspace{-2mm}
    \label{fig:finetune}
\end{figure*}

\section{Related Work}
\label{sec:relatedwork}

\noindent{\bf Image-to-image translation} aims at transferring images from a source domain to a target domain.
Its taxonomy can be generally grouped into two paradigms: supervised~\cite{isola2017image, mirza2014conditional, park2019semantic} and unsupervised~\cite{zhu2017unpaired, liu2017unsupervised, huang2018munit, park2020contrastive, jiang2020tsit}, depending on whether paired training data are available. Pix2pix~\cite{isola2017image} is the first supervised image translation model with conditional GANs~\cite{mirza2014conditional}, and is later extended to pix2pixHD~\cite{wang2018high} for generating high-resolution images. Due to the difficulty of acquiring paired images, unsupervised models have been developed, 
typically based on the assumption of the cycle-consistency constraint~\cite{zhu2017unpaired}. However,
the underlying bijective assumption is restrictive. Some methods~\cite{benaim2017one, park2020contrastive, nizan2020breaking, jiang2020tsit} have tried to break the cycle, such as the contrastive unpaired translation model (CUT)~\cite{park2020contrastive}, which uses contrastive learning to maximize the mutual information between the local patches of input and output images. 
Patches are a natural unit for learning intricate anime-style textures between images and complex relations between objects within individual images, making contrastive learning-based translations useful for scene stylization.
Thus, we build our unsupervised training branch with a recent contrastive learning-based translation model~\cite{jung2022exploring}.

\noindent{\bf Domain adaptation of StyleGAN} has been an active area of research, aiming to transfer knowledge of pre-trained GANs to new domains.
Several adaptation strategies based on fine-tuning have been proposed, including learnable parameter selection~\cite{mo2020freeze, yang2021one, wang2018transferring, robb2020few}, data augmentations~\cite{karras2020training, zhao2020differentiable, tran2021data}, and regularization terms~\cite{li2020few, ojha2021few, zhao2022closer, xiao2022few}. FreezeG~\cite{lee2020freeze} freezes the generator’s low-resolution layers to sustain the structure of the source domain. 
Recent studies~\cite{kwon2022one, wang2022ctlgan,gal2022stylegan,zhu2021mind} 
guide attribute-level adaptation by calculating the domain gap direction in a CLIP embedding space~\cite{radford2021learning}.
Despite promising attempts, StyleGAN adaptation still has limitations in fixed image resolution, failure modeling of complex scenes, overfitting, and undesired semantic artifacts. 
In comparison, these issues are well addressed by our semantic-constrained strategy and data selection, as well as the semi-supervised framework.

\noindent{\bf Scene cartoonization.}
CartoonGAN~\cite{chen2018cartoongan} proposed a semantic content loss and an edge-promoting adversarial loss to retain clear edges and smooth shading. It was further extended to lightweight AnimeGAN~\cite{chen2020animegan} with improved anime-specific loss functions. However, their global 
unsupervised learning framework is unable to capture local cartoon textures. Wang et al.~\cite{wang2020learning} introduced a white-box framework, decomposing images into surface, structure, and texture -- to guide the cartoonization process. Nevertheless, the obtained results replace fine details with flat color blocks. 
Recently, Gao et al.~\cite{gao2022learning} proposed a cartoon-texture-saliency-sampler (CTSS) module to better perceive and transfer cartoon textures. 
However, it merely learns texture abstraction and over-saturated color, limiting its ability to synthesize anime scenes with hand-drawn styles.
Different from existing efforts, our work features a novel semi-supervised image-to-image translation framework that resorts to pseudo paired data guidance to simplify this task. A patch-wise constructive loss within and between images is introduced to maintain content consistency and learn local anime textures better.


\section{Methodology}
\label{sec:method}

Our goal is to stylize natural scenes with fine-grained anime textures while preserving the underlying semantics.
We formulate the proposed \textit{Scenimefy} into a three-stage pipeline:
pseudo paired data generation (Section~\ref{sec:gen}), segmentation-guided data selection (Section~\ref{sec:selec}), and semi-supervised image-to-image translation (Section~\ref{sec:i2i}).

\subsection{Pseudo Paired Data Generation}
\label{sec:gen}
To bridge the domain gap between the real and anime scene, paired data is beneficial to establish semantic and style correspondences to ease the standard unsupervised I2I translation. 
While StyleGAN~\cite{karras2019style, karras2020analyzing} can synthesize high-quality images, the complexity 
of anime scenes necessitates an elaborately designed fine-tuning strategy for StyleGAN to generate plausible pseudo paired data.

The proposed pseudo paired data generation process is illustrated in Figure~\ref{fig:finetune}.
With a source StyleGAN $G_s$ pre-trained on the real scene dataset, we fine-tune it on the anime scene dataset to obtain $G_t$. Then, we can generate paired data $\{x^p, y^p\}$ with semantic similarity from a random latent code $w$ as $x^p=G_s(w) \in X^p$ and $y^p=G_t(w) \in Y^p$, where $X^p$ and $Y^p$ are the pseudo real scene domain and anime scene domain, respectively. Based on the observation that the early layers (\ie, the low-resolution ones) of StyleGAN determine the structure information, we freeze the initial blocks of the generator and the initial style vectors injected to preserve the spatial layout during fine-tuning.

To better preserve category-specific objects, we propose to guide $y^p$ to follow the semantic attributes of $x^p$ using pre-trained model priors, VGG~\cite{simonyan2014very} and CLIP~\cite{radford2021learning}. Specifically, we include the CLIP loss to minimize the cosine distance between the CLIP-space embeddings of the two images and use the perceptual loss~\cite{zhang2018unreasonable} to constrain the overall semantics.
\begin{equation}
\small
\label{eq:finetunegloballoss}
    L_{global} = D_{CLIP}(x^p,y^p) + \lambda_{lpip}lpips(x^p,y^p), 
\end{equation}
where $D_{CLIP}(\cdot,\cdot)$ denotes cosine distance in the CLIP space, $lpips(\cdot,\cdot)$ is the perceptual loss, and $\lambda_{lpips}$ is the loss weight. 

To better maintain local spatial information and details,
we incorporate a patch-wise contrastive loss (PatchNCE), inspired by CUT~\cite{park2020contrastive}, which applies contrastive learning to the embedded features of the generator.
We use the pre-trained CLIP models to extract feature embeddings instead of an additional MLP header network that may cause the potential imbalance issue 
during fine-tuning~\cite{kwon2022one}.
Specifically, we randomly crop patches in $x^p$ and $y^p$ and embed them with the CLIP encoder $E$, as shown in Figure~\ref{fig:finetune}.
Then, we bring the positive patches closer, which are cropped at the same position, and the negative patches far apart, which are cropped from different positions.
Let $v$ denote the embedded query patch from $y^p$. Let $v^+$ and $\{v^{-}_i\}^N_{i=1}$ be the embedded positive patch and $N$ negative patches from $x^p$, respectively.
The patch-wise loss can be written as:
\begin{equation}
\small
\label{eq:finetunepatchloss}
    L_{patch}(v,v^+,v^-) = -\text{log}\left[\frac{\text{exp}(v\cdot v^+)}{\text{exp}(v\cdot v^+) + \sum^N_{i=1}\text{exp}(v\cdot v_i^-)}\right],
\end{equation}

The overall loss function of this stage can be written as:
\begin{equation}
\small
\label{eq:overallfinetuneloss}
    L_{finetune} = L^{t}_{GAN}(G_{t},D) + \lambda_{global} L_{global} + \lambda_{patch} L_{patch},
\end{equation}
where $L^{t}_{GAN}$ is the adversarial loss~\cite{goodfellow2014generative}, and $D$ is the StyleGAN discriminator. $\lambda_{global}$ and $\lambda_{patch}$ are the loss weights.

\subsection{Semantic Segmentation Guided Data Selection}
\label{sec:selec}

\begin{figure}[!t]
    \centering
    \includegraphics[width=\linewidth]
    {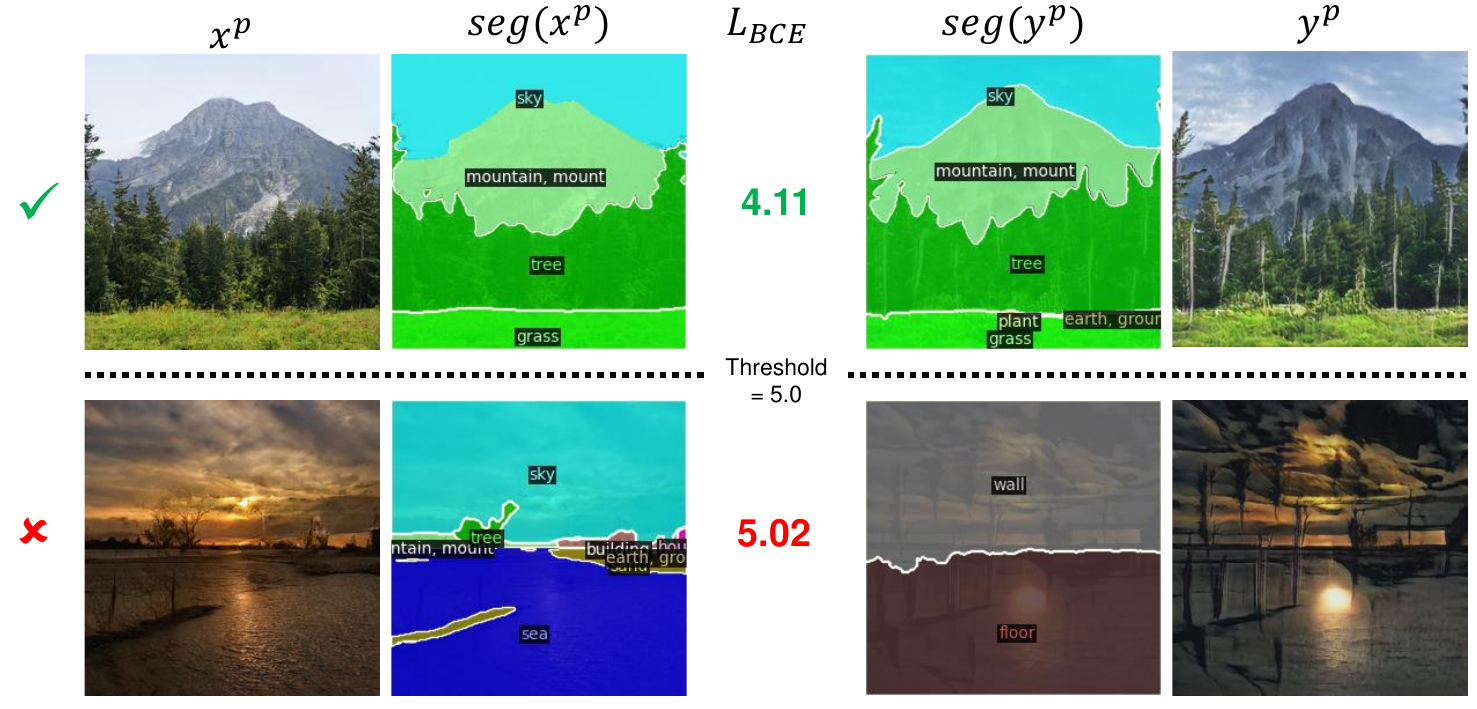}
    \vspace{-5.5mm}
    \caption{{\bf Filtering examples of the segmentation-guided data selection scheme.} We automatically filter images of low quality using the pixel-wise cross-entropy loss $L_{BCE}$. The retained pseudo paired data (top) exhibits higher structure consistency than the discarded pair (bottom).}
    \vspace{-3.5mm}
    \label{fig:semsegmask}
\end{figure}

Through pseudo paired data generation, we obtain a synthetic paired dataset with coarse pixel-wise correspondence. 
However, such raw pseudo paired data still risks low quality or poor structural consistency as shown in Figure~\ref{fig:semsegmask}, suggesting a need for data filtering.

To this end, we propose a semantic segmentation guided data selection scheme to purge low-quality samples with less structural consistency.
We observe that recent semantic segmentation models, such as Mask2Former~\cite{cheng2022masked}, can generalize well to the anime domain. Such observation allows us to use Mask2Former for pseudo paired data filtering based on two elaborately designed criteria, \ie, semantic consistency and semantic abundance.
Specifically, we employ pixel-wise cross-entropy loss $L_{BCE}$ as a metric to evaluate semantic consistency. The samples with a loss value higher than a threshold of $5.0$ are eliminated.
To enrich semantic abundance, we exclude images with only one detected category since this indicates either little semantic information or low quality.
The visualizations of the retained and filtered images and their predicted masks are presented in Figure~\ref{fig:semsegmask}, and more examples of the cleaned pseudo pairs are shown in Figure~\ref{fig:finetune}. It is observed that the remaining pseudo paired data achieves plausible quality after this stage.

\begin{figure*}[t]
    \centering
    \vspace{-1.5mm}
    \includegraphics[width=0.95\linewidth]
    {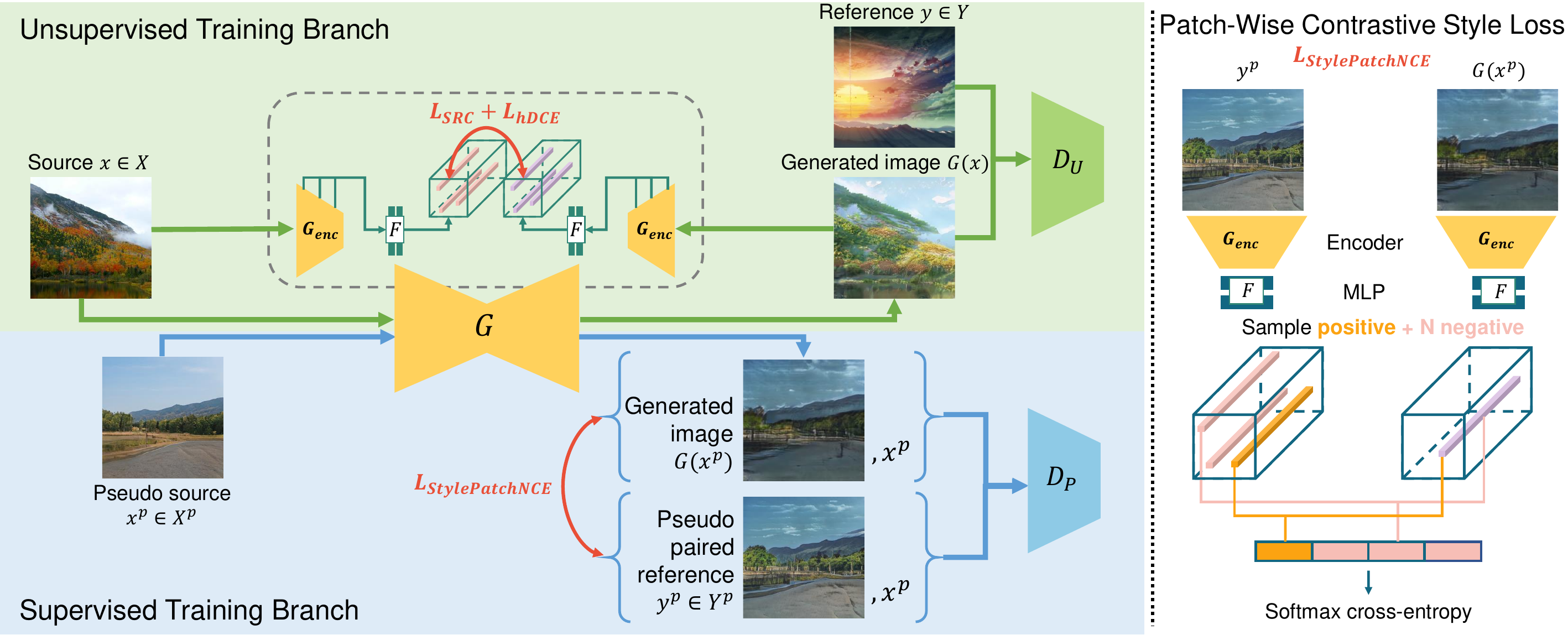}
    \vspace{-1.5mm}
    \caption{\textbf{Main framework of semi-supervised image-to-image translation.}
    \textbf{Left:} The proposed approach comprises two branches: unsupervised (top) and supervised (bottom).
    The supervised branch ingests the pseudo paired data as supervision, while the unsupervised branch learns the true target distribution using the original real and anime datasets.
    A novel patch-wise contrastive style loss is proposed to learn local fine details better.
    \textbf{Right:} Details of the patch-wise contrastive style loss.}
    \vspace{-4mm}
    \label{fig:semi-i2i}
\end{figure*}

\subsection{Semi-Supervised Image-to-Image Translation}
\label{sec:i2i}

Our semi-supervised image-to-image translation framework (see Figure~\ref{fig:semi-i2i}) consists of two branches, supervised and unsupervised. Given a set of real scene images $\{x_i\}_{i=1}^N$ in the real domain $X$ and an anime set $\{y_j\}_{j=1}^M$ in the anime domain $Y$, 
our goal is to learn a mapping $G: X \rightarrow Y$ with the help of the pseudo paired dataset $P = \{x^p_i, y^p_i\}_{i=1}^N$.
The dual-branch training procedure is detailed below.

\subsubsection{Supervised Training Branch}
\label{sec:sup}
The supervised training branch ingests the pseudo paired data to leverage the coarse pixel-wise correspondence between domain $X^p$ and $Y^p$, thus facilitating the training and semantic mapping of complex scene stylization. 
The supervised branch is based on a conditional GAN framework~\cite{mirza2014conditional}, with the conditional adversarial loss:

\begin{align}
\label{eq:cGANloss}
    \mathcal{L}_{cGAN}(G,D_P) = &\mathbb{E}_{y^p,x^p}[\log D_P(y^p,x^p)] + \nonumber \\
                 &\mathbb{E}_{x^p}[\log (1-D_P(x^p,G(x^p))],
\end{align}
where a patch discriminator $D_P$ aims to distinguish between $\{(y^p, x^p)\}$ and $\{(G(x^p), x^p)\}$, and $(\cdot,\cdot)$ denotes a concatenation operation.

Different from a typical supervised I2I framework~\cite{isola2017image} that uses a reconstruction loss to impose a strong supervision, our ground truth images lie in $Y^p$ rather than the real target $Y$. Instead, we introduce a novel patch-wise contrastive style loss for robust supervision. 
The intuition is that for a good translation, each patch within the translated image should be akin to the corresponding patch in the pseudo ground truth, rather than be identical to it.
Such patches at the same location should be embedded to be closer, whereas the ones from different locations should be far away. The patch-level contrastive learning helps our model learn robust local style similarity and focus on fine details.
We name this loss the \emph{StylePatchNCE} loss, as shown in Figure~\ref{fig:semi-i2i}.
We divide the generator $G$ into two components, the encoder $G_{enc}$ and the decoder $G_{dec}$, \ie, $G(x)=G_{dec}(G_{enc}(x))$.
The feature stack computed in $G_{enc}$ is available to conduct image translation, whose element also naturally corresponds to a patch of the input image, with deeper layers representing larger patches.
This feature is further passed through a two-layer trainable MLP network $F$, following SimCLR~\cite{chen2020simple} to obtain the embedded patch feature. 
To precisely capture anime textures at different granularity, we select multi-scale features from a total of $L$ layers of $G_{enc}$. Let $\tilde{v}_l^i$ and $v_l^i$ be the $l$-th-layer embedded patch at the location $i$ of $G(x^p)$ and $y^p$, respectively.

The proposed \emph{StylePatchNCE} loss can thus be formulated as:
\vspace{-2mm}
\begin{equation}
\small
\label{eq:10}
    L_{StylePatchNCE}(G,F,Y^p) = \sum_{l=1}^{L} \sum_{i\neq j} L_{patch}^{style}(\tilde{v}_l^i, v_l^i, v_l^j),
    \vspace{-2mm}
\end{equation}
$L_{patch}^{style}$ shares the same contrastive loss form with $L_{patch}$ in Eq.~\eqref{eq:finetunepatchloss}. The patch-level constraint enables a denser supervision of $G$, readily easing the training.

The training objective of the supervised branch is:
\begin{equation}
\small
\label{eq:supervised}
    L_{sup} = {L}_{cGAN}(G,D_P) + \lambda_{style} L_{StylePatchNCE}(G,F,Y^p),
\end{equation}
where $\lambda_{style}$ is the weight for the StylePatchNCE loss.

\subsubsection{Unsupervised Training Branch}
\label{sec:unsup}
The unsupervised branch directly ingests the original high-quality real dataset $\{x_i\}_{i=1}^N$ and anime dataset $\{y_j\}_{j=1}^M$ to learn the true target domain distribution.
Inspired by Jung~\etal~\cite{jung2022exploring}, we notice the importance of tackling heterogeneous semantic relations of the image patches within a complex scene image. For instance, the patches from a mountain or a sea, and even their different parts, have diverse semantic information. Such semantic relation should be considered and preserved for plausible unsupervised scene stylization.

Accordingly, we apply the semantic relation consistency loss $L_{SRC}$ and the hard negative contrastive loss $L_{hDCE}$ for training~\cite{jung2022exploring}. $L_{SRC}$ minimizes the Jensen-Shannon Divergence (JSD) of the in-image patch similarity distribution between $x$ and $G(x)$, to enhance semantic consistency during translation. 
$L_{hDCE}$ applies the patch-wise contrastive loss that gradually increases the discriminative difficulty of negative samples to enhance the discriminative power of the model. For the loss details, please refer to Jung~\etal~\cite{jung2022exploring}.
We apply both losses to the features of $x$ and $G(x)$ extracted by $G_{enc}$ and $F$, similar to our StylePatchNCE loss.

The total loss of the unsupervised branch is written as:
\begin{equation}
\small
\label{eq:unsup}
    L_{unsup} = {L}_{GAN}(G,D_U) + \lambda_{SRC} L_{SRC} + \lambda_{hDCE} L_{hDCE},
\end{equation}
where ${L}_{GAN}$ is the adversarial loss~\cite{goodfellow2014generative}, and $D_U$ is a standard discriminator for unsupervised training. $\lambda_{SRC}$ and $\lambda_{hDCE}$ are the loss weights.


\subsubsection{Overall Training}
\label{sec:i2iloss}
The full framework of Scenimefy is thus semi-supervised, seeking a trade-off between faithfulness and fidelity of scene stylization. The full loss function is defined as:
\begin{equation}
\small
\label{eq:i2iloss}
    L_{i2i} = L_{unsup} + \lambda_{sup}L_{sup},
\end{equation}
where $\lambda_{sup}$ decays gradually following a cosine function as the training proceeds.


\section{Experiments}
\label{sec:experiments}

\begin{figure}[t]
    \centering
    \includegraphics[width=\linewidth]
    {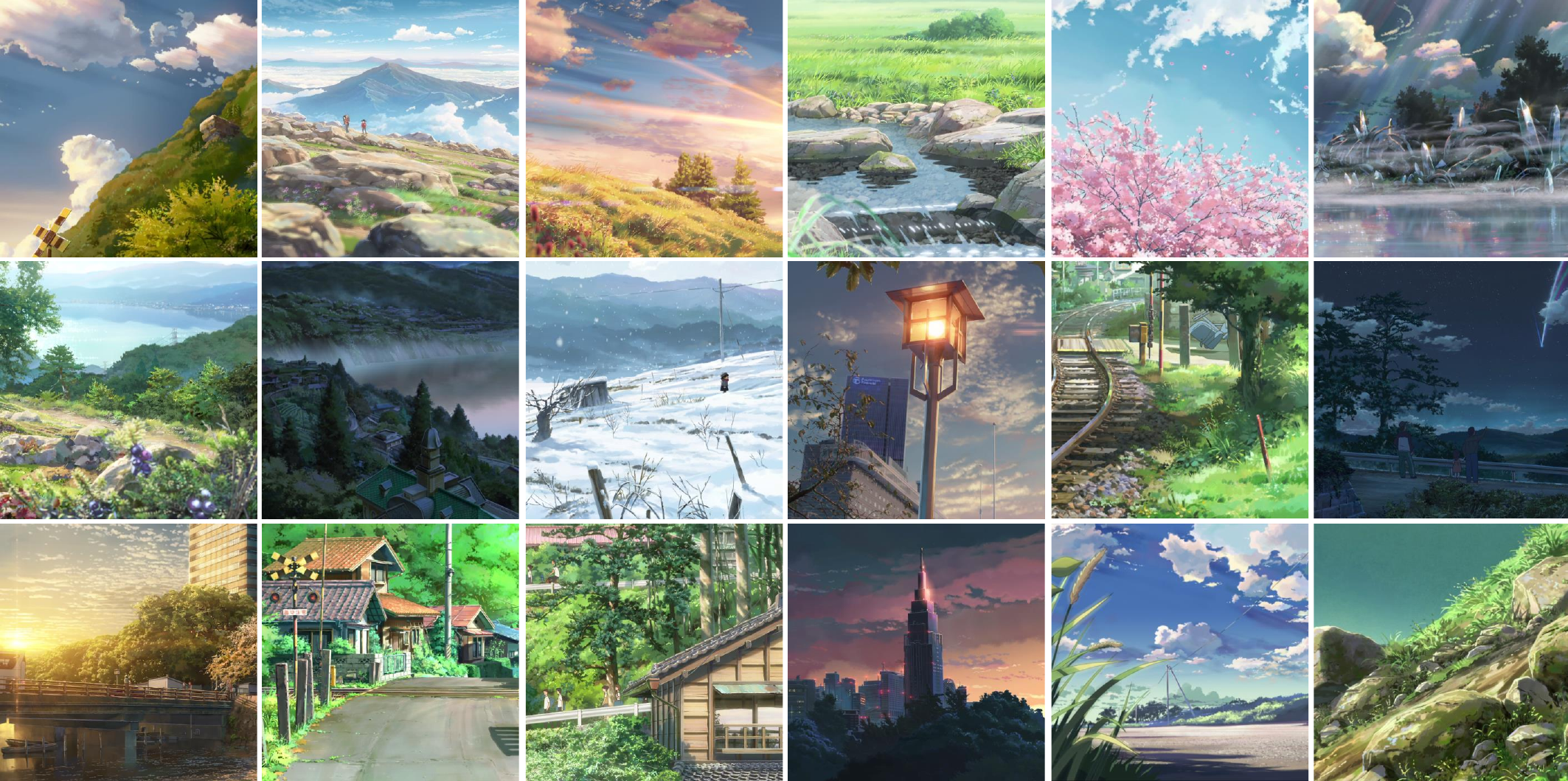}
    \caption{{\bf Examples of our anime scene dataset.} We showcase the image samples of various anime scenes collected from nine Shinkai's films.}
    \vspace{-5mm}
    \label{fig:animedataset}
\end{figure}

\begin{figure*}[!t]
    \centering
    \includegraphics[width=\linewidth]
    {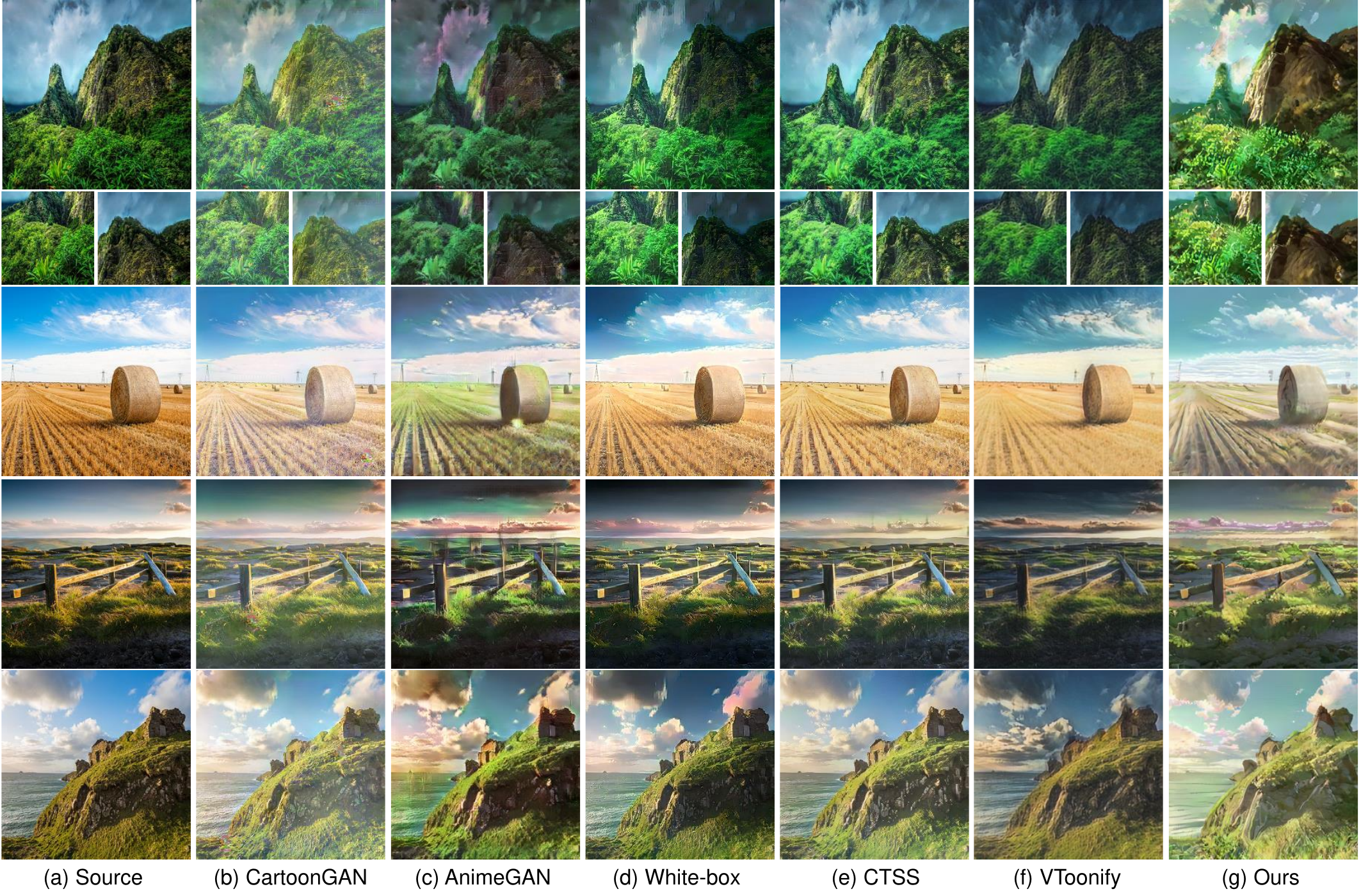}\vspace{-1mm}
    \caption{{\bf Qualitative comparison.} We compare our approach with five representative methods. Scenimefy (ours) produces more semantic-consistent results with rich anime-style textures compared to state-of-the-art baselines. Zoom in for details.}
    \label{fig:mainresults}    
\end{figure*}

\subsection{Settings}

\noindent{\bf Dataset.}
\label{sec:dataset}
As our method is a semi-supervised image-to-image translation framework, the training datasets include real-world scene photos and anime scene images for the unsupervised branch, as well as a pseudo paired dataset for the supervised branch. During training, all the images are resized to a resolution of 256$\times$256.

\emph{Real scene photos.} We used 90,000 natural landscape images from the Landscapes High-Quality (LHQ) dataset~\cite{skorokhodov2021aligning} as our training set, and 6,656 scene images provided by the authors of CycleGAN~\cite{zhu2017unpaired} as our test set.

\emph{Anime scene photos.} Our study also contributed a high-resolution (1080$\times$1080) Shinkai-style pure anime scene dataset comprising 5,958 images. To construct this dataset, we gathered key frames from nine prominent Shinkai Mokoto films, \ie, `Weathering with You' (2019), `Your Name' (2016), `Children Who Chase Lost Voices' (2011), \etc. Subsequently, we manually refined the dataset by eliminating irrelevant and low-quality images. Unlike the datasets used in previous studies~\cite{chen2020animegan, wang2020learning}, our dataset does not contain randomly cropped images with a large portion of human portraits, which exhibit notable dissimilarities in their features when compared with the background scene. This curation was done to mitigate the potential overfitting issue during fine-tuning and narrow the domain gap. Our dataset will be made publicly available to facilitate future research in scene stylization. The example images in our dataset can be found in Figure~\ref{fig:animedataset}. 

\emph{Pseudo paired dataset.} We randomly sampled 30,000 paired images with the same latent codes from the source StyleGAN generator and the fine-tuned one. We set a mild truncation trick~\cite{karras2019style} with a threshold $\gamma=0.7$ to ameliorate data quality without sacrificing much diversity.
The proposed segmentation-guided data selection (Section~\ref{sec:selec}) was applied to ameliorate data quality.

\noindent{\bf Baselines.}
We select five state-of-the-art baselines to make a comprehensive comparison. These baselines can be grouped into two categories:
1) representative image-to-image translation translation methods customized for scene cartoonization, \ie, CartoonGAN~\cite{chen2018cartoongan}, AnimeGAN~\cite{chen2020animegan}, White-box~\cite{wang2020learning}, CTSS~\cite{gao2022learning}; 
2) and the StyleGAN-based approach, \ie, VToonify~\cite{yang2022Vtoonify}.

\noindent{\bf Implementation details.} 
Regarding the training process, we first train a StyleGAN2 generator on the LHQ dataset at 256$\times$256. We then fine-tune this generator on our collected anime dataset, with the last three layers trainable and the remaining layers frozen, using the hyper-parameters $\lambda_{lpips}=0.01$, $\lambda_{global}=1.0$, $\lambda_{patch}=0.05$ for 1,000 iterations. Following this, we generate 30,000 pseudo paired data using a truncation of $0.7$ with the proposed data selection scheme described in Section~\ref{sec:selec}. Our implementation of the unsupervised training branch is based on the recent image translation model~\cite{jung2022exploring}. Scenimefy is trained on 
a single NVIDIA GeForce RTX 3090 GPU
for 20 epochs with $\lambda_{style}=0.05$, $\lambda_{SRC}=0.05$, $\lambda_{hDCE}=0.1$, $\lambda_{sup}(t)=\cos(\frac{\pi}{40}(t-1))$, where $t = \{1,2,..,20\}$ is the number of training epoch. More detailed settings are provided in the \textit{appendix}. 

\noindent{\bf Evaluation metrics.} We use Fréchet Inception Distance (FID)~\cite{heusel2017gans} to quantify the perceptual quality of translated images. 
The FID is calculated between a collection of 6,605 generated images and our introduced anime scene dataset.
A lower value indicates the better image quality.
In addition, we conduct a user study, where the candidates rate different methods in terms of stylization, semantic preservation, and overall translation quality. Higher scores signify better image quality.

\begin{table*}[!t]
\centering
\caption{\textbf{Quantitative comparison using FID.} A lower FID is better. The anime scene dataset is used as a reference.}\vspace{-1mm}
\label{tbl:fidcompare}
\begin{tabular}{cccccccc}
\hline
Method & LHQ (real)~\cite{skorokhodov2021aligning} & CartoonGAN~\cite{chen2018cartoongan} & AnimeGAN~\cite{chen2020animegan} & White-box~\cite{wang2020learning} & CTSS~\cite{gao2022learning} & VToonify~\cite{yang2022Vtoonify} & Ours \\
\hline
FID$\downarrow$ & 121.807 & 67.200 & 67.739 & 61.973 & 66.729 & 90.578 & \textbf{48.922} \\
\hline
\end{tabular}
\vspace{-2mm}
\end{table*}

\begin{table*}[!t]
\centering
\caption{\textbf{User preference scores.} The best scores are marked in bold.}\vspace{-1mm}
\label{tbl:userstudyscore}
\begin{tabular}{c|cccccc}
\hline
Method &  CartoonGAN~\cite{chen2018cartoongan} & AnimeGAN~\cite{chen2020animegan} & White-box~\cite{wang2020learning} & CTSS~\cite{gao2022learning} & VToonify~\cite{yang2022Vtoonify} & Ours \\
\hline
Style & 0.067 & 0.083 & 0.110 & 0.043 & 0.010 & \textbf{0.687} \\
Content & 0.087 & 0.080 & 0.103 & 0.123 & 0.030 & \textbf{0.577} \\
Overall & 0.073 & 0.077 & 0.103 & 0.057 & 0.017 & \textbf{0.673} \\
\hline
\end{tabular}
\vspace{-4mm}
\end{table*}

\begin{figure*}[!h]
    \centering
    \includegraphics[width=\linewidth]
    {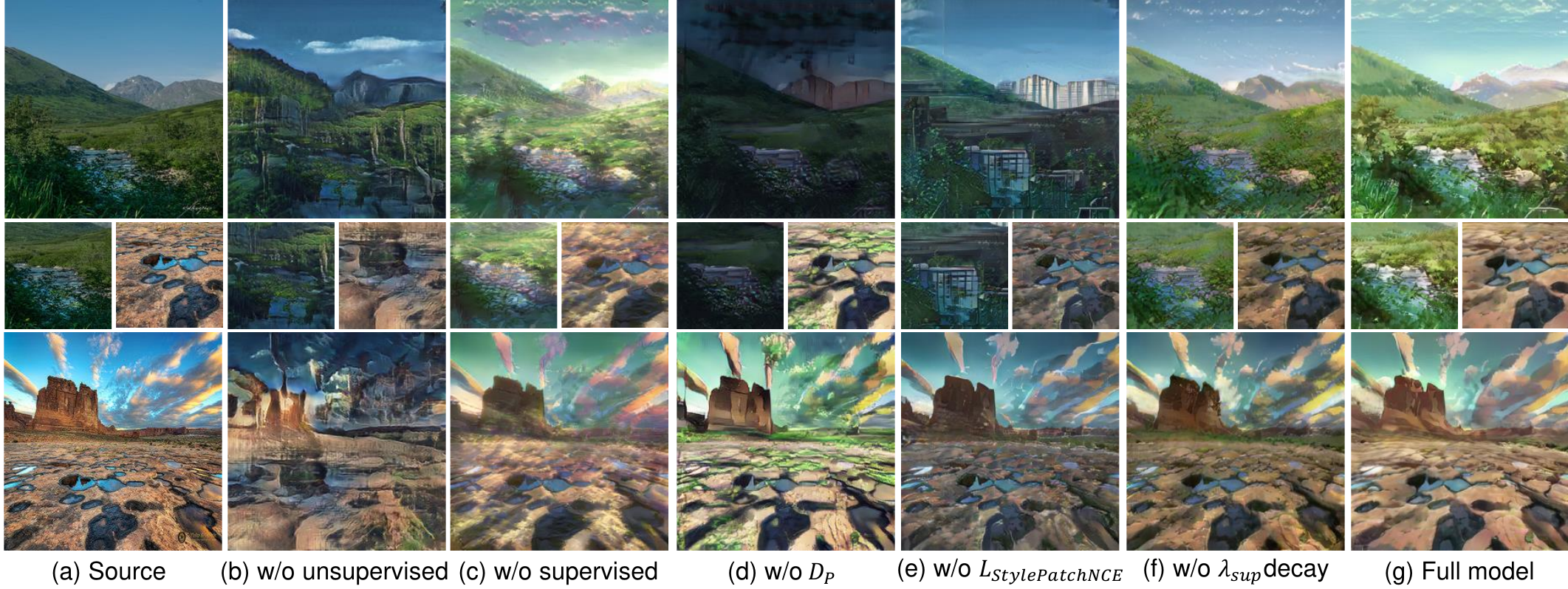}\vspace{-2mm}
    \caption{{\bf Ablation study of semi-supervised I2I translation.} The effect of each key component is illustrated.}
    \vspace{-5mm}
    \label{fig:ablationi2i}
\end{figure*}

\subsection{Main Results}

\noindent{\bf Qualitative comparison.} Figure~\ref{fig:mainresults} shows the qualitative comparison with five state-of-the-art methods. Our results seek a plausible trade-off between style fidelity and semantic faithfulness, where previous methods have either focused on pursuing more content consistency at the expense of weak anime style or have degraded into an image abstraction method, leading to the loss of fine details. CartoonGAN~\cite{chen2018cartoongan} and AnimeGAN~\cite{chen2020animegan}, design handcrafted anime-specific losses, such as the edge-smoothed loss, in attempts to manifest sharp edges, which, however, 
limits the stylization degree. In addition, 
while AnimeGAN suffers from artifacts in the sky and sea. White-box~\cite{wang2020learning} and CTSS~\cite{gao2022learning}, generate results that look more like texture abstraction with a weak style. 
When tackling scene images, the StyleGAN-based style transfer method, VToonify~\cite{yang2022Vtoonify} lacks both local anime texture and the global style like color.
All the baseline methods are unable to fully capture the inherent anime texture features.
In contrast, Scenimefy presents delicate anime features while retaining semantic consistency. For example, in the enlarged region of the first row, our model successfully mimics the brush strokes of the leaves. 
The qualitative results indicate the effectiveness of our method, outperforming the state-of-the-art baselines in perceptual quality.
More comparative results can be found in our \textit{appendix}.

\noindent{\bf Quantitative results.} In Table~\ref{tbl:fidcompare}, we conduct a quantitative evaluation of our method against the baselines.
Our approach achieves the lowest FID score, indicating that the quality of our translated results is the best, consistent with our higher visual quality.
We also test the FID between the real and the anime scene datasets as a reference.
The style distribution of our results is much closer to the anime domain compared with the real one.

In addition to FID, we conducted a user
study with 30 subjects participating to assess the quality of anime scene rendering based
on three criteria: evident anime stylization (Style), consistent semantic preservation (Content), and overall translation
performance (Overall). Participants selected what they consider to be the best
results from six different methods across 10 sets of images. 
Table~\ref{tbl:userstudyscore} summarizes the average preference scores, where
Scenimefy receives the best scores in all three criteria, thus
further suggesting the effectiveness of our method.

\begin{figure}[t]
    \centering
    \includegraphics[width=\linewidth]
    {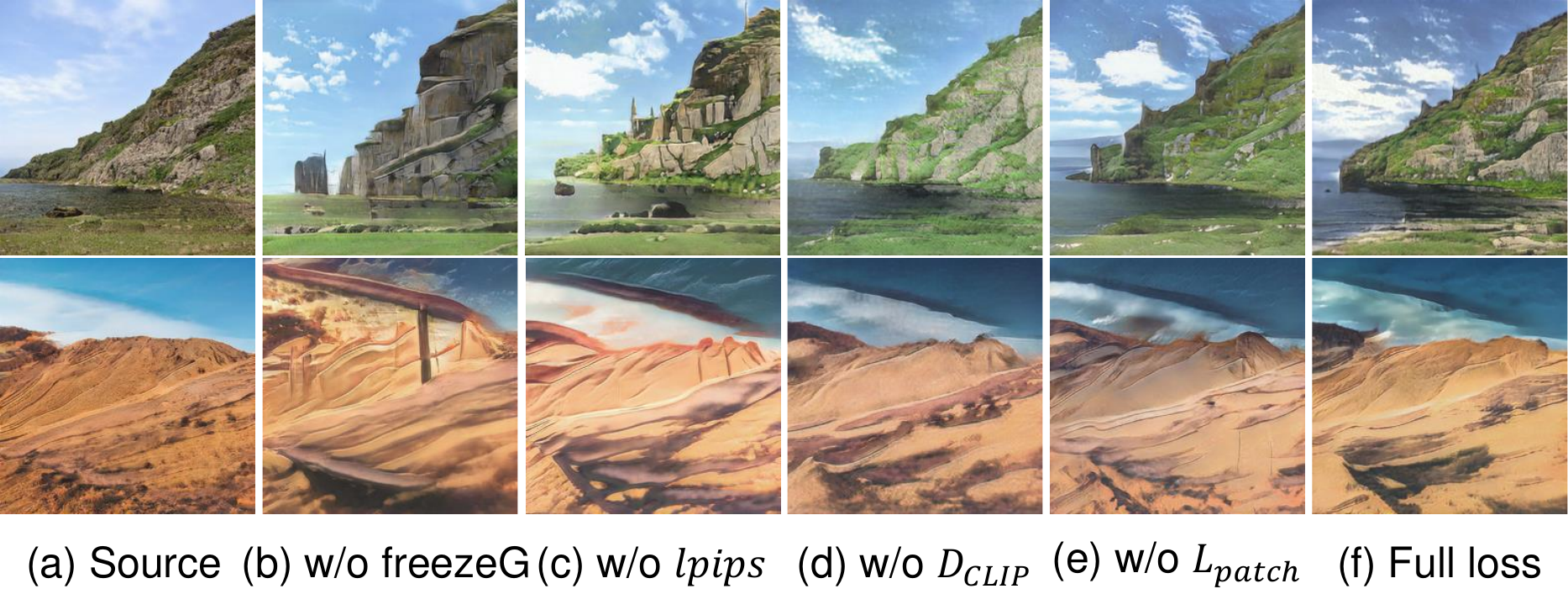}\vspace{-2mm}
    \caption{ \textbf{Ablation study of StyleGAN fine-tuning}. The influence of each key fine-tuning technique is shown.}
    \vspace{-3mm}
    \label{fig:ablfineune}
\end{figure}

\begin{table}[!h]
\centering
\caption{\textbf{$L_{BCE}$ of StyleGAN fine-tuning.} 
}
\vspace{-2mm}
\label{tbl:finetune-compare-quan}
\resizebox{\columnwidth}{!}{
    \begin{tabular}{c|cccccc}
    \hline
    Finetune & w/o freezeG & w/o \textit{lpips} & w/o $D_{CLIP}$ & w/o $L_{patch}$ & Full loss \\
    \hline
    $L_{BCE}$ $\downarrow$ & 4.45 & 4.53 & 4.25 & 4.30 & \textbf{4.24} \\
    \hline
    \end{tabular}
}
\vspace{-5mm}
\end{table}

\begin{table}[!h]
\centering
\caption{\textbf{$L_{BCE}$ of semi-supervised I2I translation.} 
}
\vspace{-2mm}
\label{tbl:semi-i2i-compare-quan}
\resizebox{\columnwidth}{!}{
    \begin{tabular}{c|cccccccc}
    \hline
    \shortstack{I2I\\\ } & \shortstack{w/o\\unsupervised} 
    & \shortstack{w/o\\supervised}
    & \shortstack{w/o\\$D_{P}$} 
    & \shortstack{w/o\\$L_{StylePatchNCE}$}
    & \shortstack{w/o\\$\lambda_{sup}$decay}
    & \shortstack{Full\\model} \\
    \hline
    $L_{BCE}$ $\downarrow$ & 4.76 & 4.50 & 4.68 & 4.72 & 4.58 & \textbf{4.39} \\
    \hline
    \end{tabular}
}
\vspace{-5mm}
\end{table}

\begin{figure*}[!t]
    \centering
    \vspace{-5mm}
    \includegraphics[width=\linewidth]
    {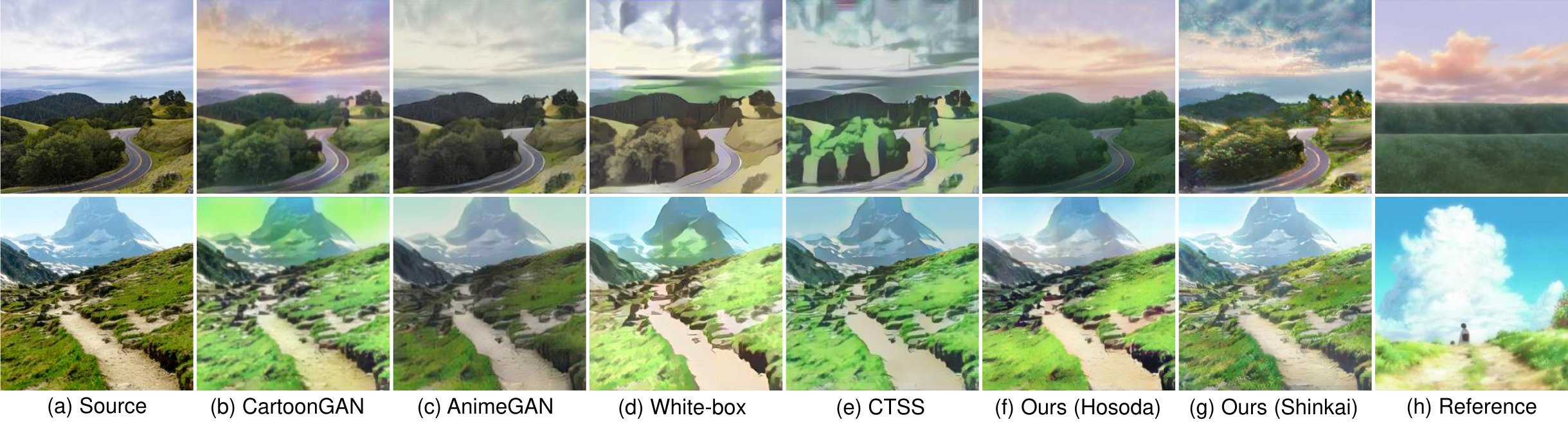}
    \vspace{-7mm}
    \caption{{\bf Hosoda anime style.} We compare our approach with four baseline methods on the Hosoda dataset from \cite{wang2020learning}. Scenimefy (ours) exhibits a better ability to capture different anime-style textures. Zoom in for details.}
    \label{fig:hosoda}
    \vspace{-4mm}
\end{figure*}

\subsection{Ablation Study}

\noindent{\bf StyleGAN fine-tuning.} The effect of each semantic-constrained technique for fine-tuning is shown in Figure~\ref{fig:ablfineune}. Without freezing the shallow layers of the pre-trained StyleGAN (see Figure~\ref{fig:ablfineune}(b)), the spatial structure is severely altered. 
Ablating the global constraints ($lpips$ or $D_{CLIP}$) from the pre-trained model prior, category-specific objects are poorly preserved, such as odd rock textures on mountains, as shown in Figure~\ref{fig:ablfineune}(c)(d).
The removal of the patch-wise consistency loss in Figure~\ref{fig:ablfineune}(e) results in a loss of fine details.
Applying all the pre-trained model prior constraints at both image-level and patch-level, our full method accurately generates valid pairs, transfers the anime style while maintaining the semantic structure, and has fewer artifacts.  

\noindent{\bf Semi-supervised image-to-image translation.}\label{sec:ablai2i} To verify the efficacy of the proposed semi-supervised image-to-image translation framework, we conduct a systematic ablation study by removing each key module independently. The visual results are shown in Figure~\ref{fig:ablationi2i}. 
It is observed that training each branch separately results in poor outputs.
Employing the supervised branch alone (Figure~\ref{fig:ablationi2i}(b)) leads to low-quality results due to coarse guidance.
Meanwhile, although the unsupervised branch alone successfully learns global anime style, it lacks evident local texture stylization and semantic preservation, and suffers notable artifacts, \eg, translating the water into stones in Figure~\ref{fig:ablationi2i}(c). 
Without the conditional discriminator, the results exhibit discernible local details, as depicted in Figure~\ref{fig:ablationi2i}(d). 
Ablating the patch-wise contrastive style loss leads to noisy and weird patterns on the mountain in Figure~\ref{fig:ablationi2i}(e). 
The coarse pseudo data may lead to negative effects, such as less evident style effects in Figure~\ref{fig:ablationi2i}(f). We tackle this problem by a weight decay technique of the supervised branch. The results of our full model, as shown in Figure~\ref{fig:ablationi2i}(g), exhibit superior anime rendering ability, including anime texture details, harmonious colors, and much less noise. All the modules work together to improve the overall performance of the proposed method. 

\noindent{\bf{Quantitative comparison.}} We applied  $L_{BCE}$ metric (detailed in Section~\ref{sec:selec}, lower is better) to our ablation studies
for a more comprehensive quantitative evaluation on the semantic consistency (see Table~\ref{tbl:finetune-compare-quan} and ~\ref{tbl:semi-i2i-compare-quan}). 
In the StyleGAN fine-tuning experiments, we generate $3,000$ images with the same seeds and calculated their respective $L_{BCE}$ loss against the nature images. 
For I2I translation, our test dataset comprises $6,656$ images from ~\cite{zhu2017unpaired}.
We achieved the best scores in both experiments, verifying the effectiveness of our design in better semantic preservation. 


\subsection{Further Analysis}
\noindent{\bf Other anime styles.}
We trained our model over a different anime dataset, \ie, the Hosoda Mamoru dataset, comprising $5,107$ images from \cite{wang2020learning}, to validate its versatility. 
This dataset includes similar scenes cropped from the same movie frame, along with numerous human portraits, resulting in a significant domain gap. Thus, we used a global perceptual loss~\cite{zhang2018unreasonable} between the input and output images to help maintain content consistency.
The qualitative results, illustrated in Figure~\ref{fig:hosoda}, showcase the efficacy of our method in learning and transferring desired anime styles. Notably, the grass exhibits distinct styles. In the Hosoda style, the grass texture appears more simplistic (Figure~\ref{fig:hosoda}(f)), while in the Shinkai style, it appears more detailed and colorful (Figure~\ref{fig:hosoda}(g)).

\begin{figure}[!t]
    \centering
    \vspace{-2mm}
    \includegraphics[width=\linewidth]
    {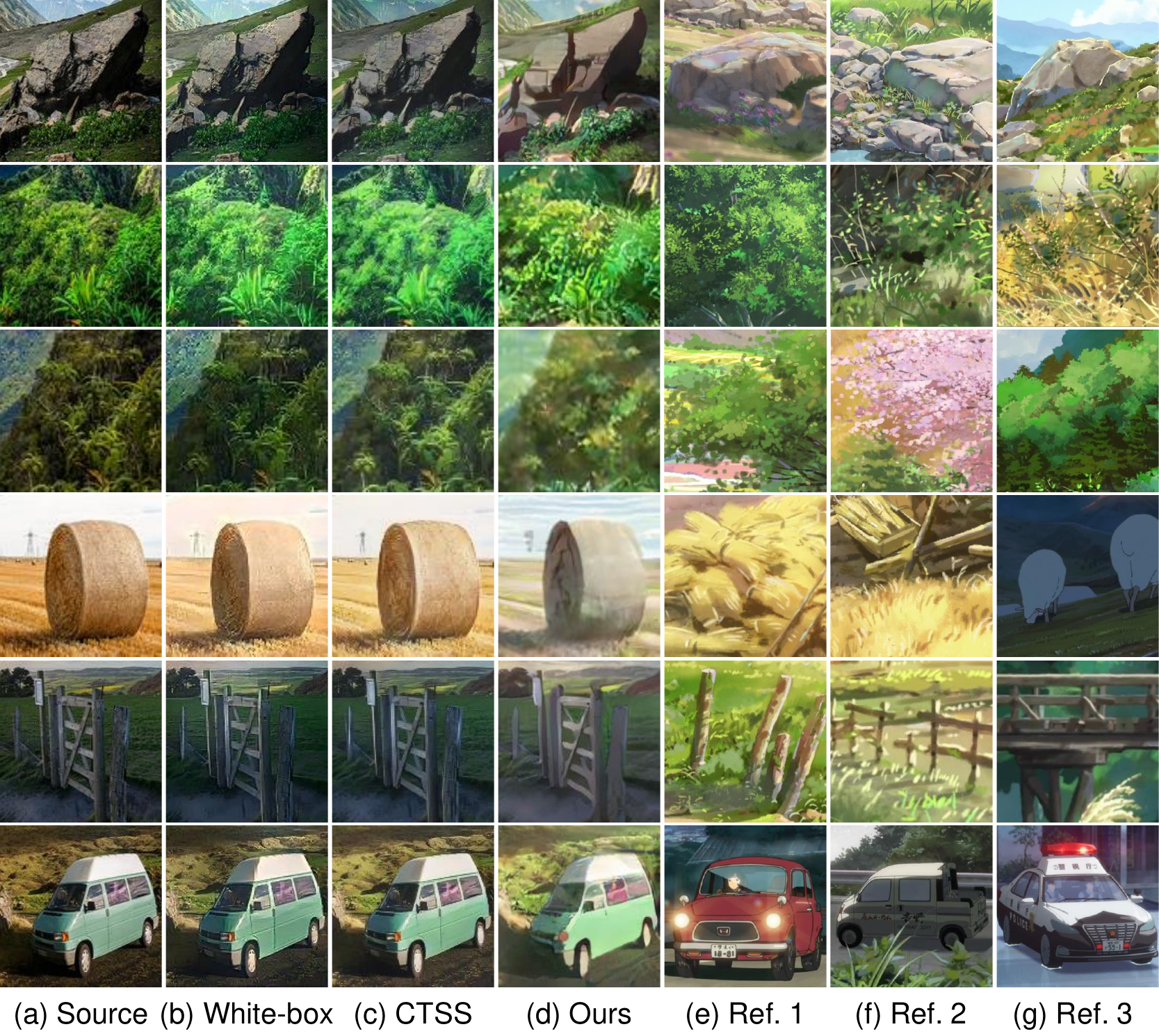}
    \vspace{-6mm}
    \caption{{\bf Detailed anime texture transfer comparison.} 
    The texture details between real anime and the generated ones by different methods.}
    \label{fig:texture-compare}
    \vspace{-7mm}
\end{figure}

\noindent{\bf Anime texture transfer.}
\label{sec:animetexture}
As in the real anime images in Figure~\ref{fig:texture-compare}(e-g), the rock, hay bale, car and background plants (Row 3) are smooth, and only
the foreground plants (Row 2) are detailed. The fence has
curvy edge and smooth textures (Row 5). Our model is the
only one that adheres to these key characteristics of anime
scenes, while other baselines overemphasize the fidelity to
the original real images so that their results look less like real
anime.

\begin{figure}[!h]
    \centering
    \vspace{-2mm}
    \includegraphics[width=\linewidth]
    {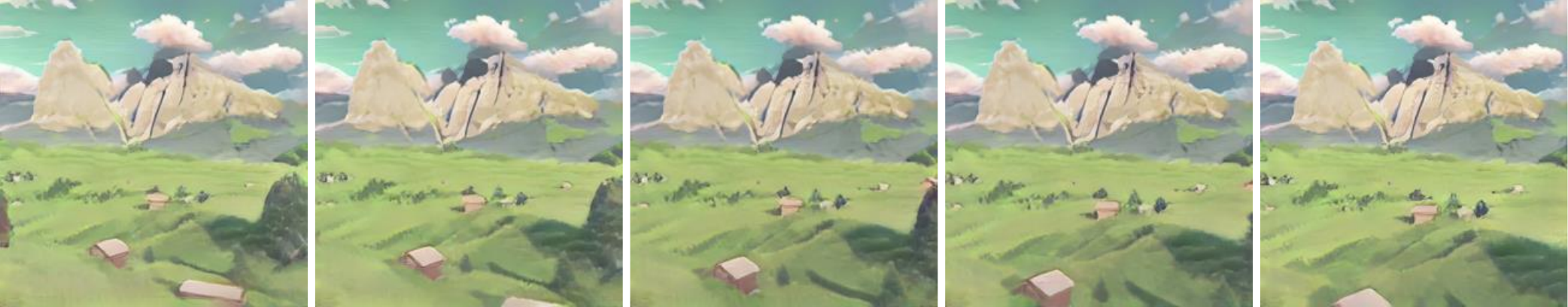}
    \vspace{-5mm}
    \caption{{\bf Temporal consistency on videos.} The results of representative frames in video stylization.}
    \label{fig:temporal-video} 
    \vspace{-2mm}
\end{figure}

\noindent{\bf Temporal consistency on videos.}
We simply extend our method to video stylization by transferring each single frame. 
The results of some representative frames are shown in Figure~\ref{fig:temporal-video}, where Scenimefy
maintains smooth and coherent visual information.

\vspace{-5mm}

\section{Conclusion}
\label{sec:conclusion}

In this paper, we proposed Scenimefy, a powerful framework for anime scene rendering, which comprises three stages: pseudo paired data generation, semantic segmentation-guided data selection, and semi-supervised image-to-image translation.
Besides, we contributed a high-resolution anime scene dataset to facilitate future research in scene stylization.
Our results empirically demonstrated that the use of soft pseudo paired data guidance can effectively balance style fidelity and semantic faithfulness, simplifying the pure unsupervised setting. The proposed contrastive style loss facilitates fine detail generation. Scenimefy thus outperforms state-of-the-art baselines in both perceptual quality and quantitative performance.
Despite promising results, there remain a few exciting avenues for improvement, such as incorporating explicit control of stylization degree and enabling more flexible translations with user-input style. 
Recent breakthroughs in diffusion models have enabled remarkable image generation capability. By leveraging these advancements, we can obtain improved pseudo-paired data with enhanced details. We believe that harnessing the potential of large-scale text-to-image models may further elevate the quality of automatic anime scene rendering.

\noindent{\bf Acknowledgment.} This study is supported under the RIE2020 Industry Alignment Fund Industry Collaboration Projects (IAF-ICP) Funding Initiative, as well as cash and in-kind contribution from the industry partner(s). It is also supported by Singapore MOE AcRF Tier 2 (MOE-T2EP20221-0001) and the NTU NAP Grant.

{\small
\bibliographystyle{ieee_fullname}
\bibliography{egbib}
}

\clearpage
\section*{Appendix}
\label{sec:appendix}

The document provides supplementary information that is not elaborated on in our main paper due to the space constraints: implementation details (Section~\ref{sec:implement}), additional comparative results (Section~\ref{sec:addcomp}), loss variant study (Section~\ref{sec:addablation}), more results of Scenimefy (Section~\ref{sec:moreresults}), and potential limitations (Section~\ref{sec:limitation}). 
{Code: \url{https://github.com/Yuxinn-J/Scenimefy}}.

\appendix
\section{Implementation Details}
\label{sec:implement}


\subsection{Network Architecture}
\noindent{\bf StyleGAN finetuing.} 
For the StyleGAN2 finetuning implementation, we modified the code based on FreezeG~\cite{lee2020freeze} by incorporating a freezed style vector and low-resolution layers of the generator.
Only the last $3$ generator blocks were made trainable, each comprising $2$ style blocks and a ToRGB layer. 
To maintain global consistency using pre-trained model priors, we utilized the VGG-19~\cite{simonyan2014very} based perceptual loss~\cite{zhang2018unreasonable} to extract features of $G_s(w)$ and $G_t(w)$ up to the $conv4\_4$ layer. For the pre-trained CLIP~\cite{radford2021learning} model, the official version of ViT-B/32 was applied. 
For the patch-wise contrastive loss, We selected $16$ patches of $32\times32$ at random locations from the images generated by $G_s(w)$ and $G_t(w)$, followed by embedding the image patches using the same CLIP encoder.

\noindent{\bf Semi-supervised image-to-image (I2I) translation.}
To implement the semi-supervised I2I translation, we adopted the official source code of~\cite{jung2022exploring} in a single-modal setting. Our generator architecture is based on CUT~\cite{park2020contrastive}, using the ResNet-based generator~\cite{he2016deep} with $9$ residual blocks for training. It contains $2$ downsampling blocks, $9$ residual blocks~\cite{he2016deep}, and $2$ upsampling blocks. A ResNet block is a conv block with skip connections, including a convolution, normalization, ReLU, convolution, normalization, and a residual connection. The downsampling and upsampling blocks contain a two-stride convolution/deconvolution, normalization, and ReLU. The full unsupervised branch is mainly built upon~\cite{jung2022exploring}.

To incorporate the supervised branch with our proposed \emph{StylePatchNCE} loss, we utilized the same PatchGAN discriminator~\cite{isola2017image} architecture as our conditional discriminator, classifying whether $70\times70$ overlapped patches are real or fake.

Regarding the architecture of the generator and layers used for the \emph{StylePatchNCE} loss, we define the first half of the generator $G$ and a 2-layer MLP as an encoder, which is represented as $G_{enc}$ and $F$. 
To calculate our multi-layer patch-wise contrastive style loss, we extract features from a total of 5 layers of $G_{enc}$, which are RGB
pixels, the first and second downsampling convolution, and the first and the fifth residual block, corresponding to layer ids $0, 4, 8, 12, 16$. 
These layers correspond to receptive fields of sizes $1\times1$, $9\times9$, $15\times15$, $35\times35$, and $99\times99$. 
Following CUT, for each layer's features, we sampled $256$ patches from random locations and applied $F$ to obtain $256$-dimensional final features.

\subsection{More Details on Data Selection and Training}
We presented most of the data selection and model training details in our main paper. Below we show some additional details.

\noindent{\bf Semantic segmentation guided data selection.}
To filter the data, the Mask2Former~\cite{cheng2022masked} semantic segmentation model with Swin-B (IN21k) as the backbone and pre-trained on the ADE20K~\cite{zhou2017scene} dataset was utilized.

\noindent{\bf More training details.} The training of our semi-supervised I2I translation model uses a single NVIDIA GeForce RTX $3090$ GPU and a batch size of one. At each iteration, the generator $G$ forwards twice, generating both $G(x)$ and $G(x^p)$. After calculating the total loss, backpropagation is  performed once. The conditional discriminator $D_P$ is trained to distinguish between the concatenated $\{(y^p, x^p)\}$ and $\{(G(x^p), x^p)\}$. During training, the model consumes approximately $5$k MiB memory. In terms of the inference speed, the rate is tested as $0.027$ second per image, achieving real-time anime-style rendering.

\section{Additional Comparative Results}
\label{sec:addcomp}

\begin{figure*}[]
    \centering
    \includegraphics[width=\linewidth]
    {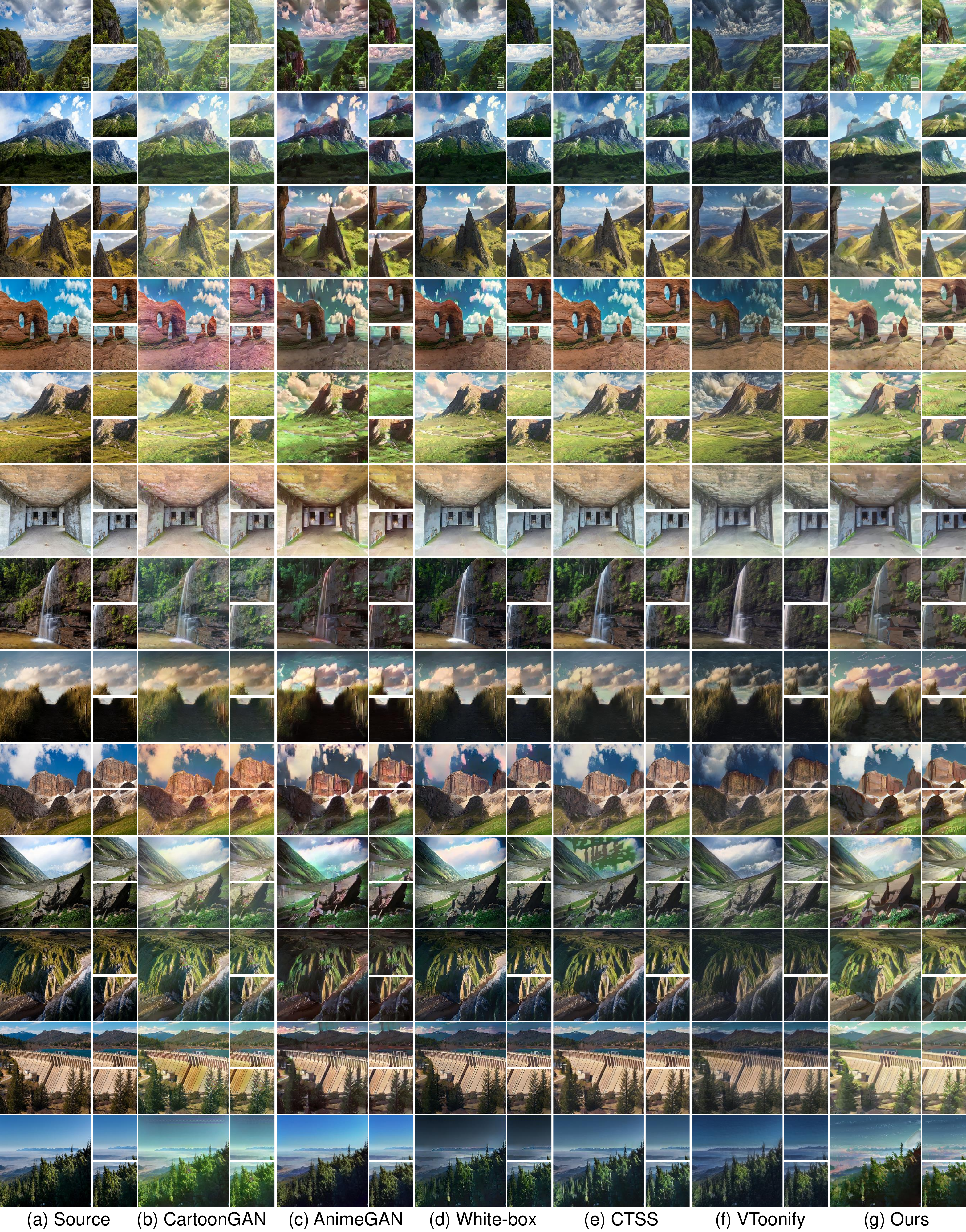}
    \caption{{\bf Additional qualitative comparative results.} 
    Zoom in for details.}
    \vspace{-2mm}
    \label{fig:suppmainresultss2} 
\end{figure*}

\begin{figure*}[]
    \centering
    \vspace{-7mm}
    \includegraphics[width=0.98\linewidth]
    {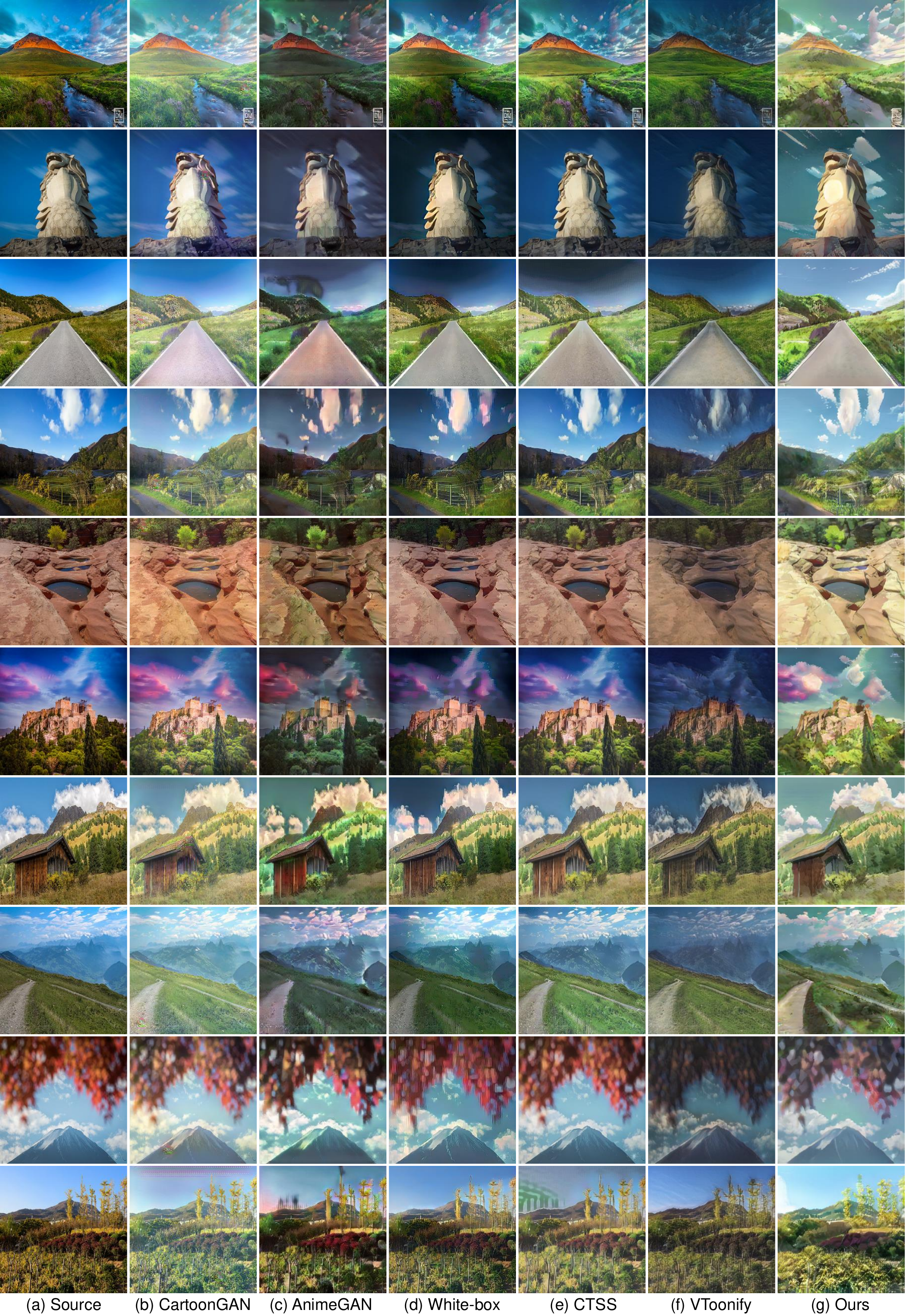}\vspace{-2mm}
    \caption{{\bf Additional qualitative comparative results.} 
    Zoom in for details.}
    \label{fig:suppmainresults} 
\end{figure*}

\begin{figure*}[]
    \centering
    \vspace{-7mm}
    \includegraphics[width=0.98\linewidth]
    {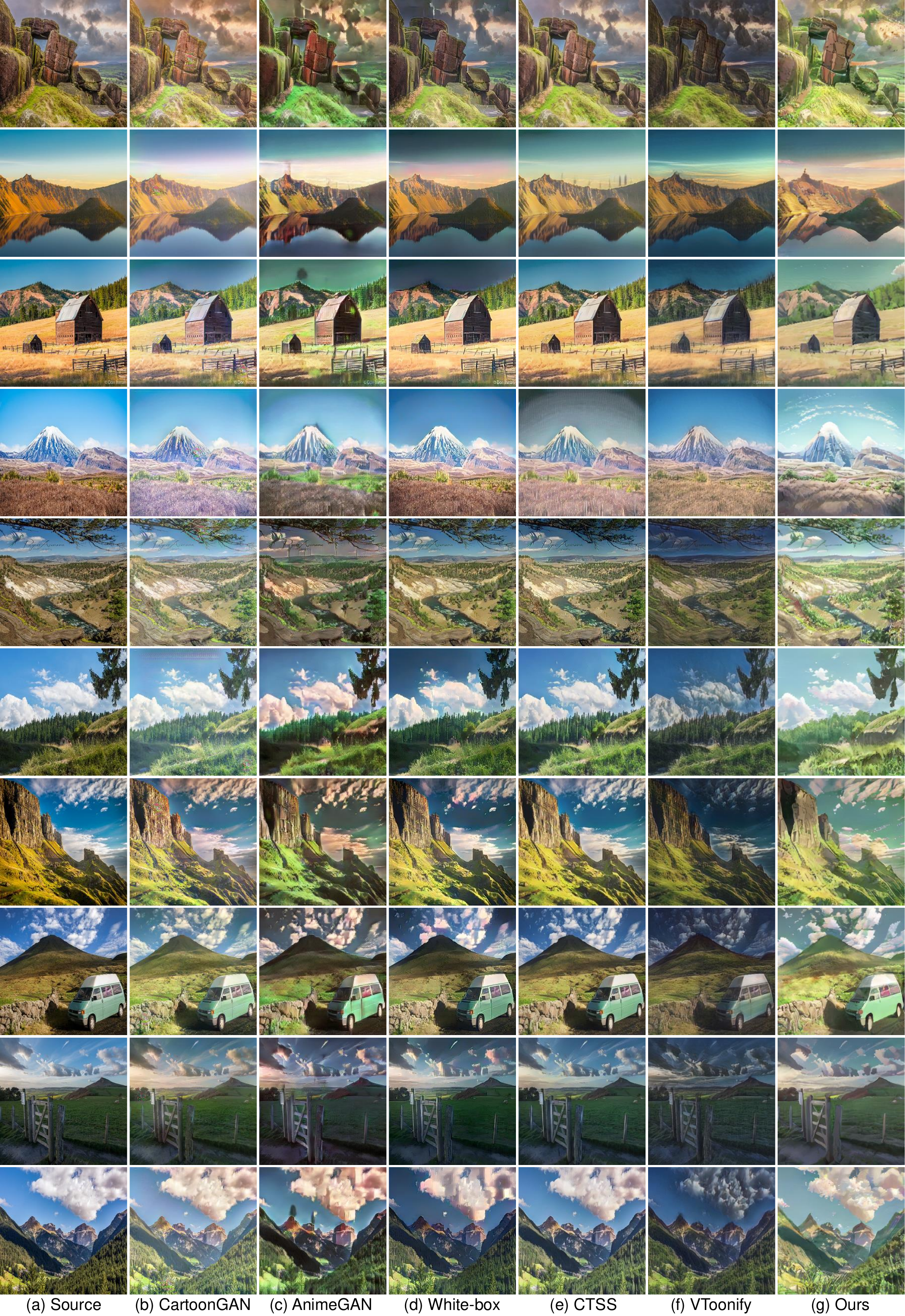}\vspace{-2mm}
    \caption{{\bf Additional qualitative comparative results.} 
    Zoom in for details.}
    \label{fig:suppmainresults3} 
\end{figure*}

\noindent{\bf More visual comparisons.}
In Section $4.2$ of the main paper, we compared our methods with five state-of-the-art methods. Due to space limit, four groups of examples were presented.
In Figures~\ref{fig:suppmainresultss2}, \ref{fig:suppmainresults} and \ref{fig:suppmainresults3}, we provide additional comparative results to further demonstrate the exceptional performance of our approach. All the previous baselines generate inferior results than our method in terms of evident stylization, consistent semantic preservation, and fine-detailed anime textures. Scenimefy achieves a more faithful anime stylization with richer colors and stronger artistic expression, resulting in the best quality output.

\section{Loss Variant Study}
\label{sec:addablation}

As mentioned in Section $3.3.1$ of the main paper, we introduce a novel patch-wise contrastive style loss, \ie, \emph{StylePatchNCE}, instead of the strict reconstruction loss~\cite{isola2017image} used in typical supervised I2I translation frameworks. We have ablated the proposed \emph{StylePatchNCE} loss in Section $4.3$ of the main paper to show its usefulness.
To further verify its effectiveness, here we directly replace the \emph{StylePatchNCE} loss with the standard $L_1$ reconstruction loss as a variant for a more comprehensive comparison.

\begin{figure}[!t]
    \centering
    \includegraphics[width=\linewidth]
    {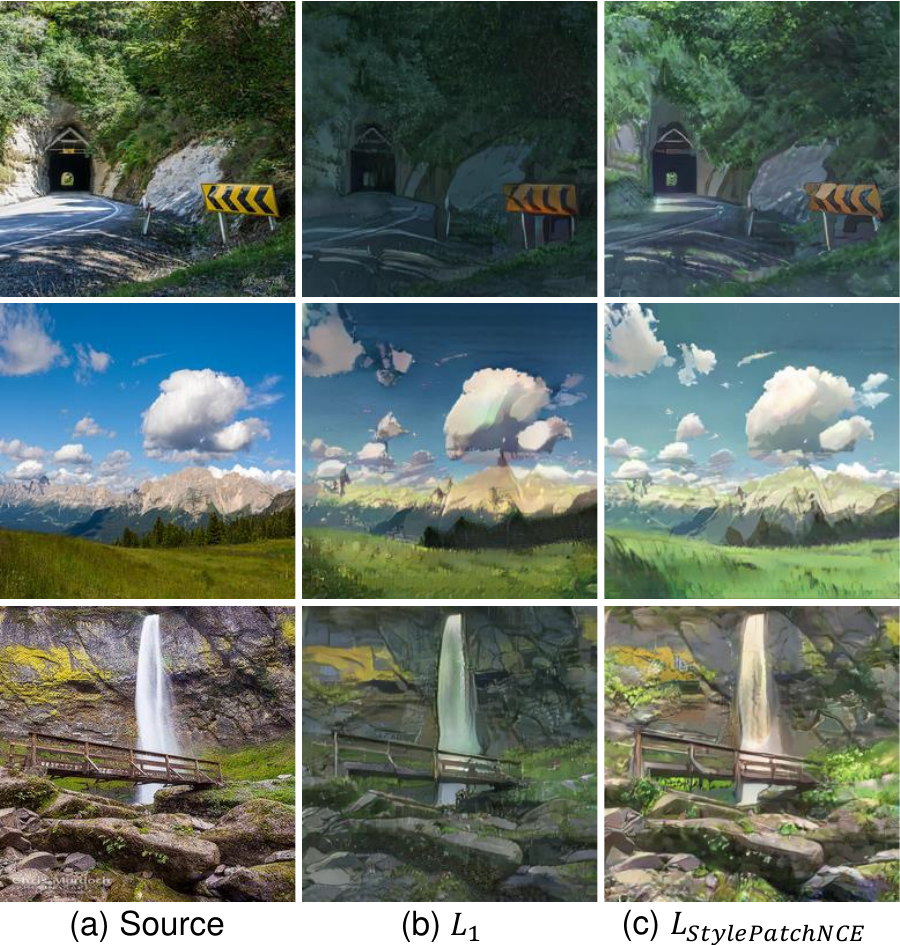}\vspace{-2mm}
    \caption{{\bf Effects of the loss function variants for the supervised branch.} We compare the effects of the $L_1$ loss with $L_{StylePatchNCE}$ loss applied to pseudo paired data.}
    \label{fig:lossvariant} 
\end{figure}

The qualitative results are presented in Figure~\ref{fig:lossvariant}. The $L_1$ loss variant resulted in an unnatural color and learned poor local anime textures, such as the clouds (in Figure~\ref{fig:lossvariant}(b), Row $2$) and the stones (in Figure~\ref{fig:lossvariant}(b), Row $3$). In contrast, our method successfully translated both the global anime style and local anime textures.

\section{More Examples Generated by Scenimefy}
\label{sec:moreresults}
\subsection{Additional Results on Scene Stylization}
\begin{figure*}[]
    \centering
    \vspace{-8mm}
    \includegraphics[width=0.94\linewidth]
    {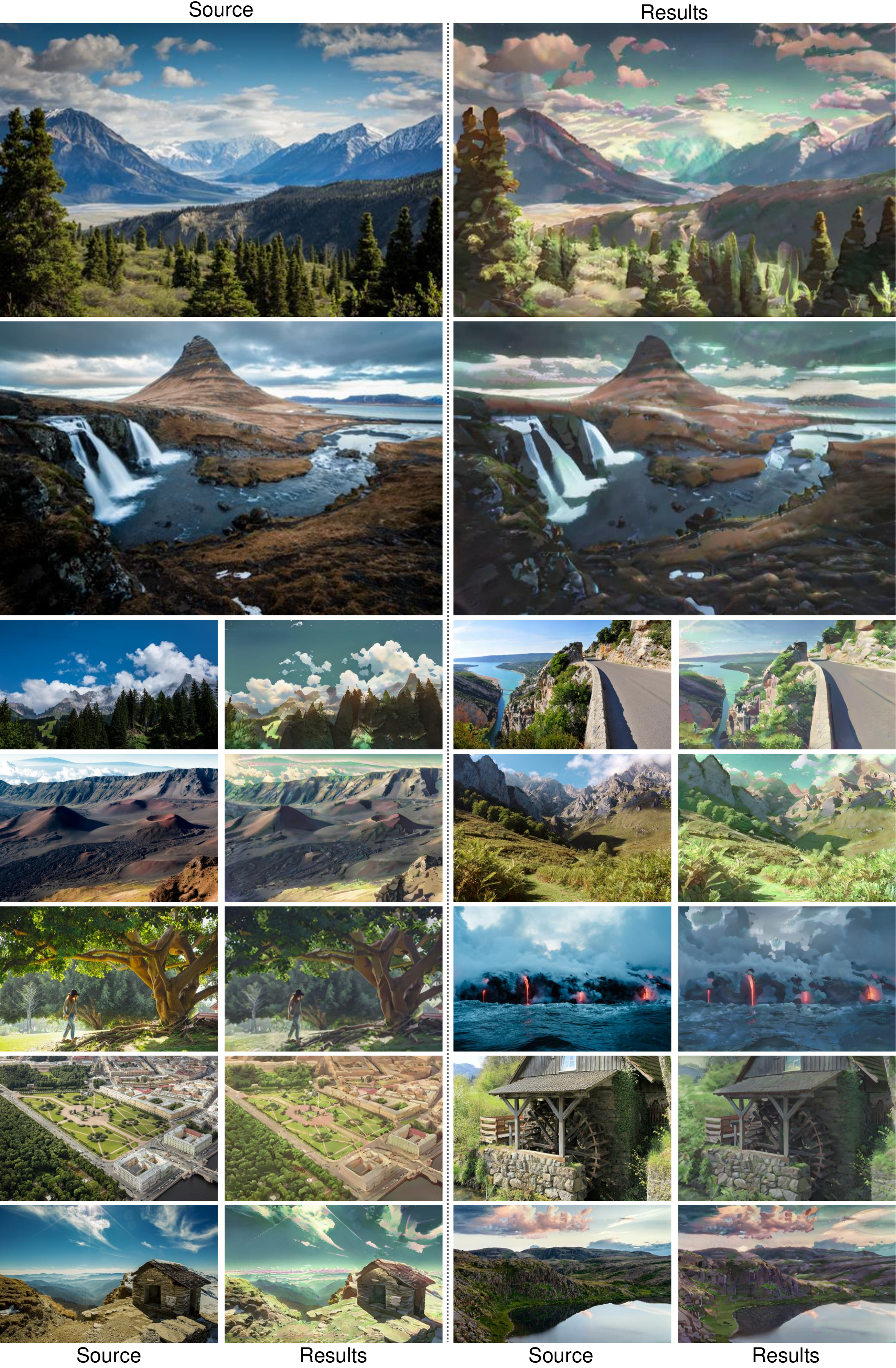}\vspace{-2mm}
    \caption{{\bf Additional results of Scenimefy on natural landscapes.} Zoom in for details.}
    \label{fig:naturelandscape} 
\end{figure*}

\begin{figure*}[t]
    \centering
    \includegraphics[width=\linewidth]
    {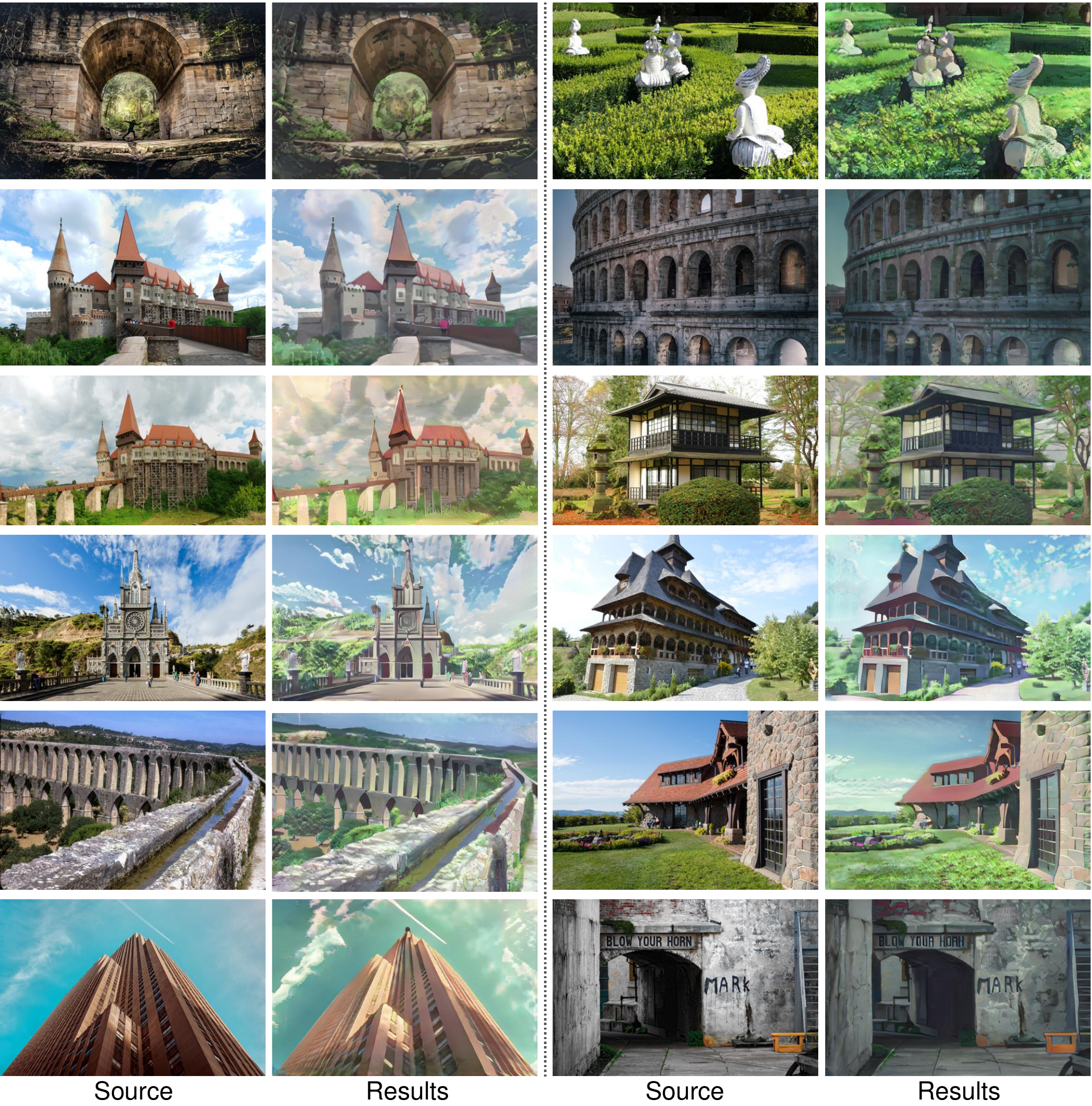}
    \caption{{\bf Additional results of Scenimefy on architecture and buildings.} Zoom in for details.}
    \label{fig:building} 
\end{figure*}

We present more generated images (in Figures~\ref{fig:naturelandscape} and \ref{fig:building}) by Scenimefy. Thanks to the fully convolutional neural network, our model can produce high-quality rectangular images despite being trained on squared images with a resolution of 256$\times$256.

\subsection{Generalizability to Other Cases}

\begin{figure*}[t]
    \centering
    \vspace{-8mm}
    \includegraphics[width=\linewidth]
    {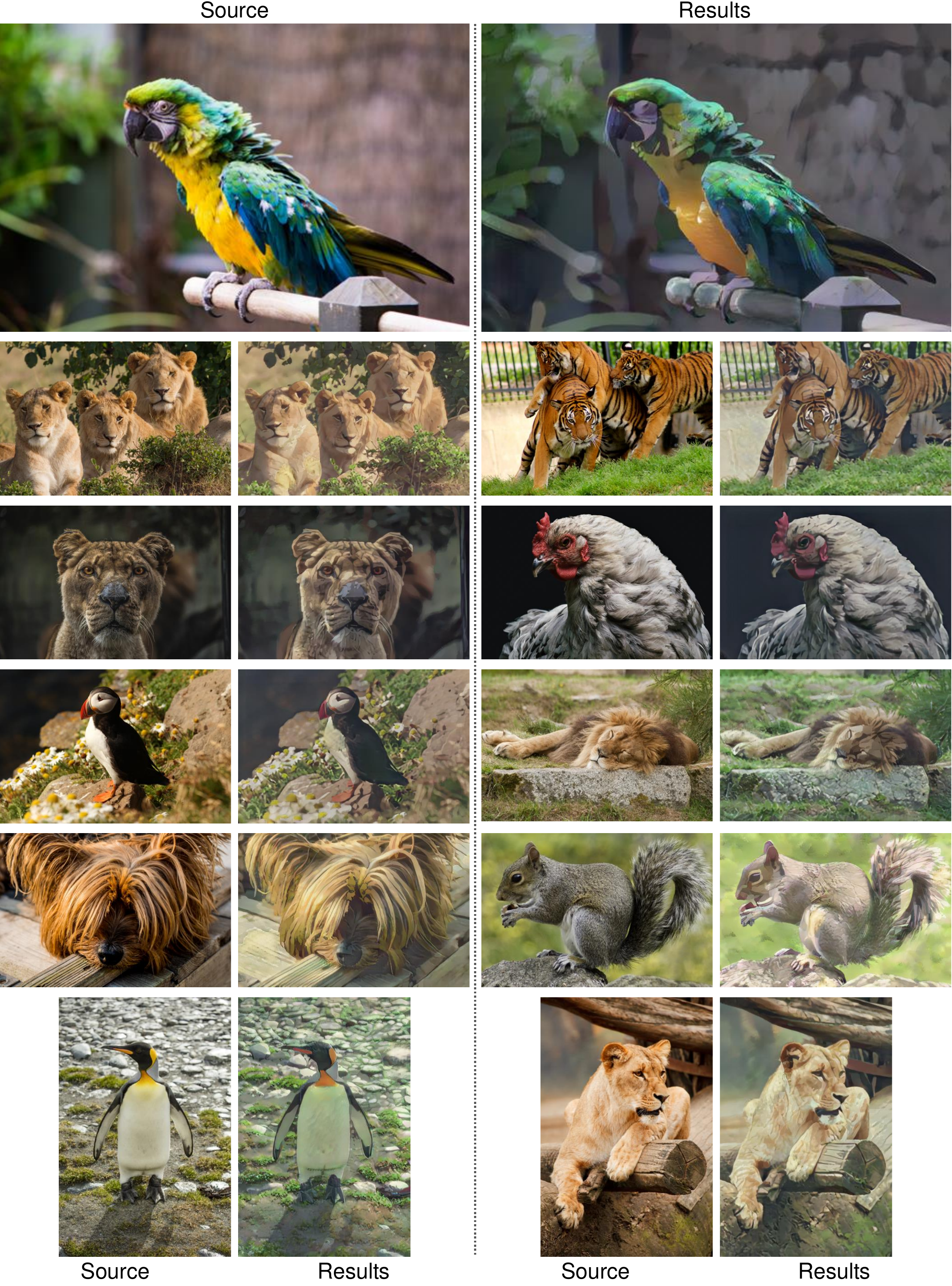}\vspace{-2mm}
    \caption{{\bf Additional results of Scenimefy on animals.} Zoom in for details.}
    \label{fig:animals} 
\end{figure*}

\begin{figure*}[t]
    \centering
    \includegraphics[width=\linewidth]
    {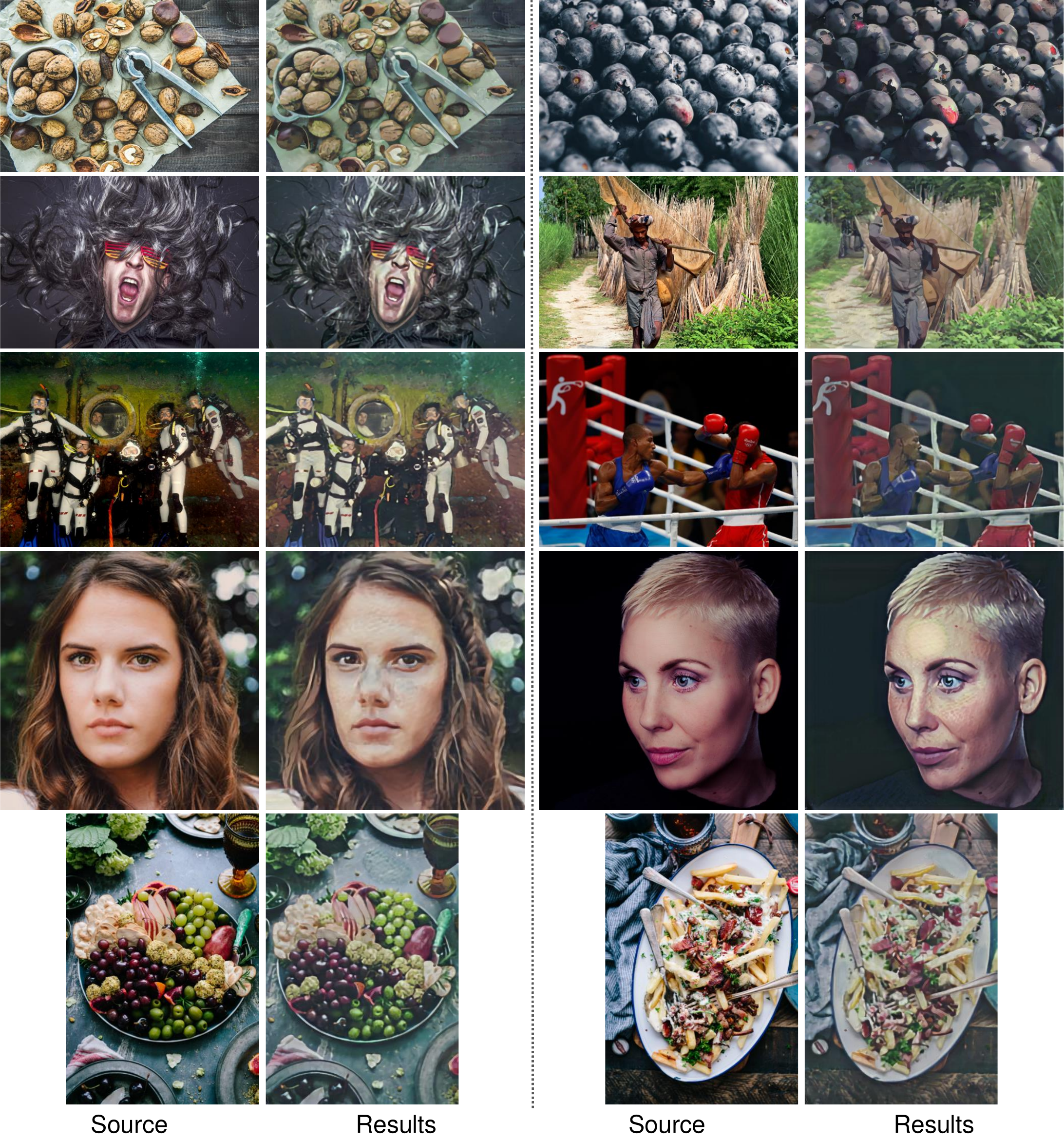}
    \caption{{\bf Additional results of Scenimefy on food and people.} Zoom in for details.}
    \label{fig:food-people} 
\end{figure*}

\begin{figure*}[t]
    \centering
    \includegraphics[width=\linewidth]
    {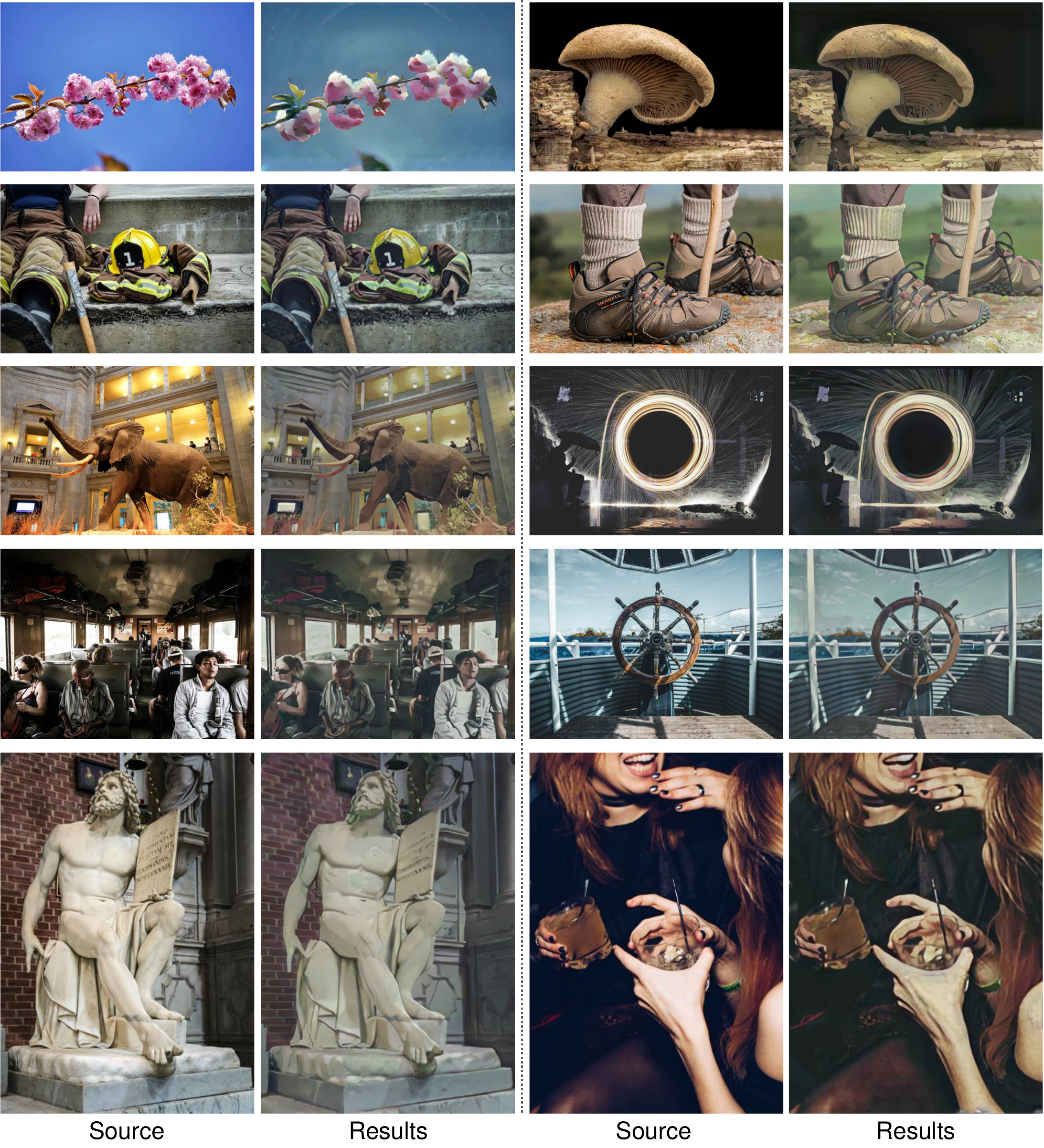}
    \caption{{\bf Additional results of Scenimefy on other objects.} Zoom in for details.}
    \label{fig:other} 
\end{figure*}

When we directly apply our model trained on natural scene landscapes~\cite{skorokhodov2021aligning}, we find that it performs relatively well even in some other cases.
The content of the test data includes city views, architecture, portraits, animals, plants, food, and other objects from the DIV2K~\cite{Agustsson_2017_CVPR_Workshops} dataset. Overall, the results (see Figures~\ref{fig:animals}, \ref{fig:food-people} and \ref{fig:other}) demonstrate that our method can generate high-quality anime stylized images in other diverse use cases and real-world scenes, indicating its certain generalizability.

\begin{figure}[!t]
    \centering
    \includegraphics[width=\linewidth]
    {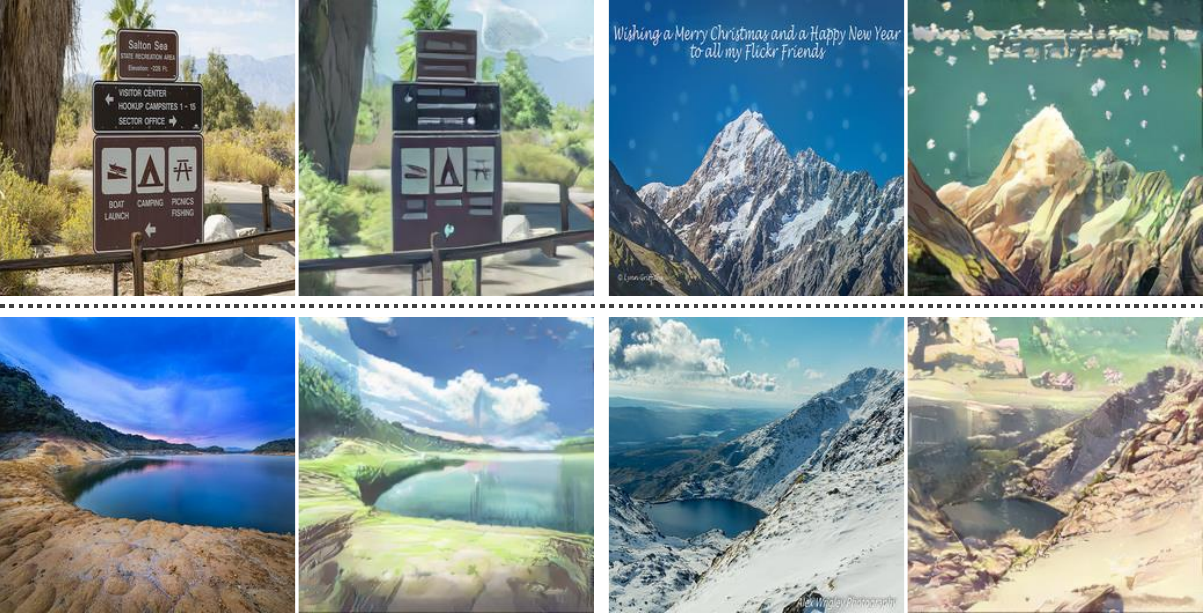}\vspace{-2mm}
    \caption{{\bf Certain failure cases.} Top: the relatively poor preservation of the tiny text-like details; Bottom: the incorrect semantic translation in a small number of cases.}
    \label{fig:limitation} 
    \vspace{-2mm}
\end{figure}

\section{Limitations}
\label{sec:limitation}
Despite the promising results, our method also has certain limitations.
Due to the absence of strong constraints imposed between the output image and the source image, our model fails to preserve intricate tiny details, such as text, as depicted in Figure~\ref{fig:limitation} (Top). However, this limitation may be overcome by introducing a simple content loss, such as the reconstruction loss or perceptual loss. This encourages the model to preserve the content of the source image more closely, albeit at the cost of slightly compromising the anime-style effect.

Moreover, our model exhibits a small number of failure cases, where it translates semantically distinct objects incorrectly due to the scene complexity and the biases in the training data. For instance, as shown in the bottom row of Figure~\ref{fig:limitation}, the model translates the stone ground into the grass, as well as the ice mountain into the stone mountain. Addressing these failure cases with more semantic consistency can also be interesting future work apart from the ones we discussed in Section $5$ of the main paper.

\end{document}